\documentclass[11pt]{article}

\usepackage[preprint]{acl}

\usepackage{times}
\usepackage{latexsym}

\usepackage[T1]{fontenc}

\usepackage[utf8]{inputenc}

\usepackage{microtype}

\usepackage{inconsolata}

\usepackage{graphicx}
\usepackage{placeins}

\usepackage[utf8]{inputenc}
\usepackage[T1]{fontenc}
\usepackage{algorithm}
\usepackage{algorithmic}
\usepackage{graphicx}
\usepackage{booktabs}
\usepackage{multirow}
\usepackage{amsmath}
\usepackage{amsfonts}
\usepackage{enumitem}
\usepackage{url} 
\usepackage{xspace}
\usepackage[table]{xcolor}
\usepackage{colortbl}
\newcolumntype{g}{>{\columncolor[rgb]{.886,.937,.855}}c}
\usepackage{soul} 
\usepackage{hyperref}
\hypersetup{
  colorlinks=true,
  urlcolor=blue
}
\usepackage{epigraph}
\usepackage{amsthm}
\usepackage{amssymb}
\usepackage{arydshln}
\usepackage{bbm}
\usepackage{wrapfig}
\usepackage{subcaption}
\usepackage{siunitx}     
\usepackage{pifont}
\usepackage{longtable}
\usepackage{booktabs}
\usepackage{array}
\usepackage{multirow}
\usepackage{pdfpages}
\usepackage{longtable}  
\usepackage{listings}   
\usepackage{textcomp}   
\usepackage{cprotect}
\usepackage[most]{tcolorbox}
\usepackage{fvextra}

\usepackage{listings}

\newcommand{\gain}[1]{\textcolor{green!60!black}{\tiny\bfseries (+#1)}}
\newcommand{\loss}[1]{\textcolor{red!80!black}{\tiny\bfseries (-#1)}}

\theoremstyle{plain}

\theoremstyle{definition}

\theoremstyle{remark}

\makeatletter
\AtBeginDocument{%
  \let\oldref\ref
  \renewcommand{\ref}[1]{%
    \hyperref[{#1}]{\underline{\oldref{#1}}}%
  }%
}
\makeatother

\usepackage{titletoc}
\newcommand\DoToC{%
  \startcontents
  \printcontents{}{1}{\textbf{\large Contents of Appendix}\vskip3pt\hrule\vskip5pt}
  \vskip3pt\hrule\vskip5pt
}

\title{SupChain-Bench: Benchmarking Large Language Models for \\ Real-World Supply Chain Management}

\author{
  \textbf{Shengyue Guan}\textsuperscript{1} \quad \textbf{Yihao Liu}\textsuperscript{2} \quad \textbf{Lang Cao}\textsuperscript{3}
  \\
  \textsuperscript{1}Alibaba Group \quad \textsuperscript{2}Peking University \quad \textsuperscript{3}University of Illinois Urbana-Champaign
  \\
  \texttt{guanshengyue.gsy@alibaba-inc.com, haoeliu@stu.pku.edu.cn, langcao2@illinois.edu} 
}

\begin{document}
\maketitle
\begin{abstract}
Large language models (LLMs) have shown promise in complex reasoning and tool-based decision making, motivating their application to real-world supply chain management. However, supply chain workflows require reliable long-horizon, multi-step orchestration grounded in domain-specific procedures, which remains challenging for current models. To systematically evaluate LLM performance in this setting, we introduce SupChain-Bench, a unified real-world benchmark that assesses both supply chain domain knowledge and long-horizon tool-based orchestration grounded in standard operating procedures (SOPs). Our experiments reveal substantial gaps in execution reliability across models. We further propose SupChain-ReAct, an SOP-free framework that autonomously synthesizes executable procedures for tool use, achieving the strongest and most consistent tool-calling performance. Our work establishes a principled benchmark for studying reliable long-horizon orchestration in real-world operational settings and highlights significant room for improvement in LLM-based supply chain agents. Our code is available at \url{https://github.com/Damon-GSY/SC-bench}.
\end{abstract}

\section{Introduction}

\begin{figure}[t]
\centering
\includegraphics[width=0.95\columnwidth]{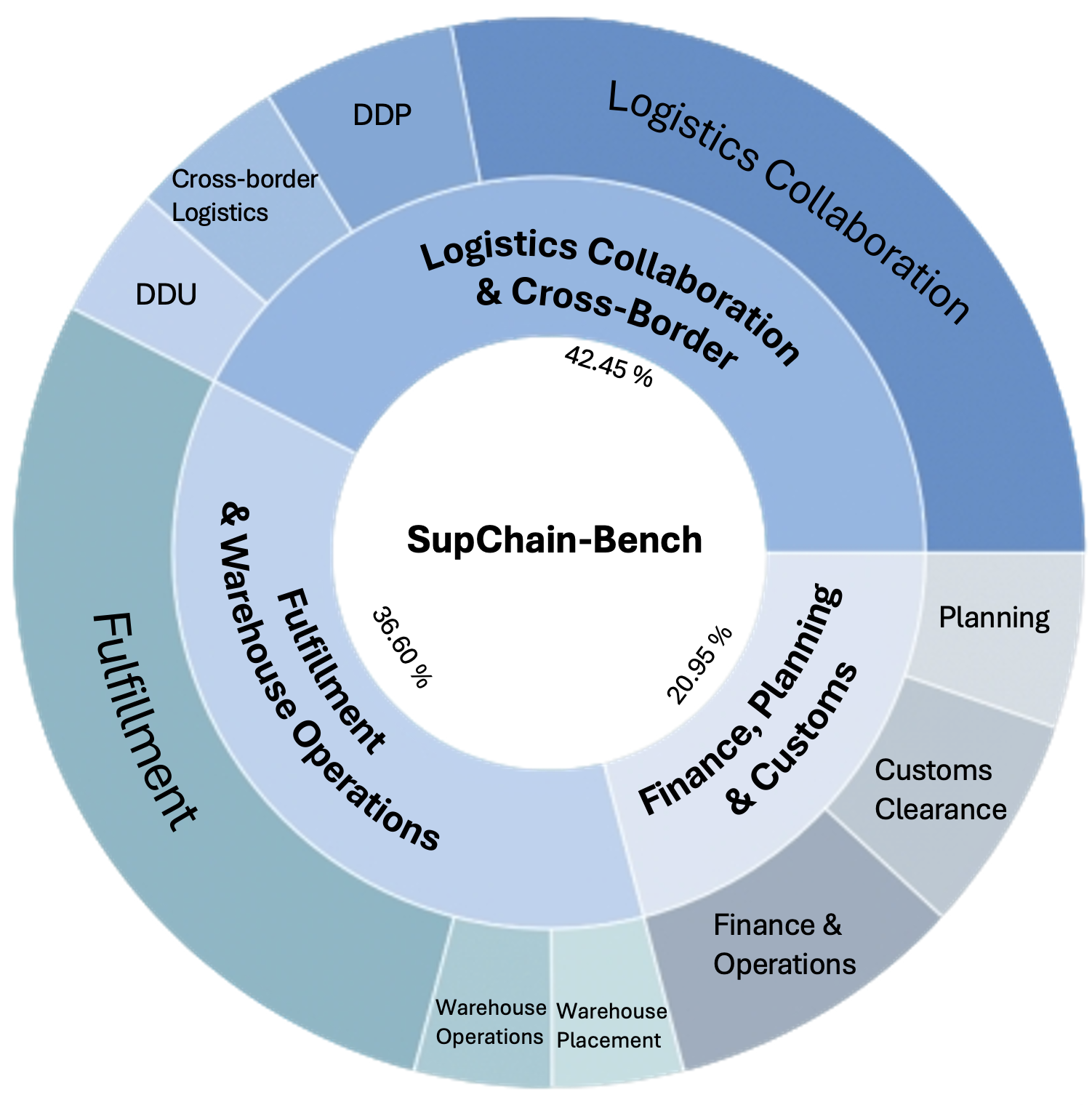}
\caption{Overall composition of SupChain-Bench. The figure shows the distribution of annotated samples across three major functional domains of supply chain management: Logistics Collaboration \& Cross-Border, Fulfillment \& Warehouse Operations, and Finance, Planning \& Customs. Each domain is further decomposed into its constituent sub-tasks, highlighting the relative proportions of different operational activities represented in the dataset.}
\label{fig:dataset-pie}
\end{figure}


The rapid evolution of Large Language Models (LLMs) has sparked a paradigm shift in their ability to support complex reasoning \citep{li202512surveyreasoning, openai_introducing_gpt5_2025_en}, strategic planning \citep{wei-etal-2025-plangenllms, yang2025qwen3technicalreport}, and autonomous agent orchestration \citep{luo2025largelanguagemodelagent}. Within this broader landscape, Supply Chain Management (SCM) emerges as a critical application domain that demands high-level intelligence to coordinate sourcing, production, and logistics under conditions of extreme uncertainty.

Supply chain management is inherently complex, driven by the global scale of operations, evolving customer demands, and the increasing need for agile, data-informed decision making. A central challenge in SCM lies in the effective integration of fragmented and often disconnected systems, spanning inventory management and demand forecasting to logistics and transportation networks. These systems must contend with demand volatility \citep{kumar2025inventory}, fragmented data landscapes \citep{akbar2024smart}, and coordination failures across organizational boundaries \citep{felder2025smart,Chen2025Optimized}, leading to operational inefficiencies, execution errors, and missed opportunities for optimization. Traditional SCM solutions, which rely heavily on rigid, rule-based logic and handcrafted heuristics, struggle to adapt to the dynamic, multi-faceted nature of modern supply chains. As a result, effective decision support increasingly demands models capable of fusing heterogeneous signals and performing multi-step reasoning over complex operational constraints, rather than optimizing isolated components in isolation.


This complexity has long motivated the use of machine learning (ML) methods to address specific SCM sub-tasks, such as demand forecasting and inventory optimization \citep{kumar2025inventory}. However, these approaches are typically designed for isolated components and lack the flexible reasoning and adaptivity required in dynamic, end-to-end supply chain environments. Recent advances in LLMs have therefore attracted growing interest as a potential mechanism for translating high-level human intent into actionable decisions and coordinated execution across heterogeneous systems. Despite this promise, the absence of a standardized and comprehensive evaluation framework remains a major bottleneck in assessing whether LLMs can reliably bridge intent and execution in practice.

To bridge this gap, we introduce a novel, domain-specific benchmark designed to assess LLM capabilities in supply chain orchestration. The overall composition of the dataset is illustrated in Figure~\ref{fig:dataset-pie}. Our benchmark addresses the challenge of fragmented evaluation by jointly assessing knowledge understanding and practical execution capabilities. Specifically, our methodology incorporates two key components: a \textbf{\textit{(1) Knowledge QA module}} that evaluates domain expertise under both context-rich and context-poor settings to assess information synthesis; and a \textbf{\textit{(2) Tool-Calling mechanism}} that challenges models to execute long-horizon, multi-step workflows. Crucially, we further introduce a comparative evaluation of Standard Operating Procedures (SOPs), enabling a principled distinction between protocol adherence and autonomous problem solving.

The main contributions of this work are summarized as follows:
\begin{itemize}[leftmargin=*, itemsep=0pt, labelsep=5pt, topsep=0pt]
    \item We present the first unified benchmark for supply chain management that jointly evaluates domain knowledge and end-to-end operational orchestration, spanning scenarios from order cancellation to logistics execution.
    \item We provide a granular analysis of LLM performance in tool-use settings, with a particular focus on quantifying the impact of SOP guidance on execution reliability and reasoning consistency.
    \item We propose SupChain-ReAct, an SOP-free framework that autonomously synthesizes executable procedures for tool use, achieving the strongest and most consistent tool-calling performance across models.
    \item We establish a strong baseline for future research by identifying concrete limitations of current LLMs in long-horizon planning and domain-specific constraint handling.
\end{itemize}


\section{Related Work}

\begin{figure*}[!t]
    \centering
    \includegraphics[width=\linewidth]{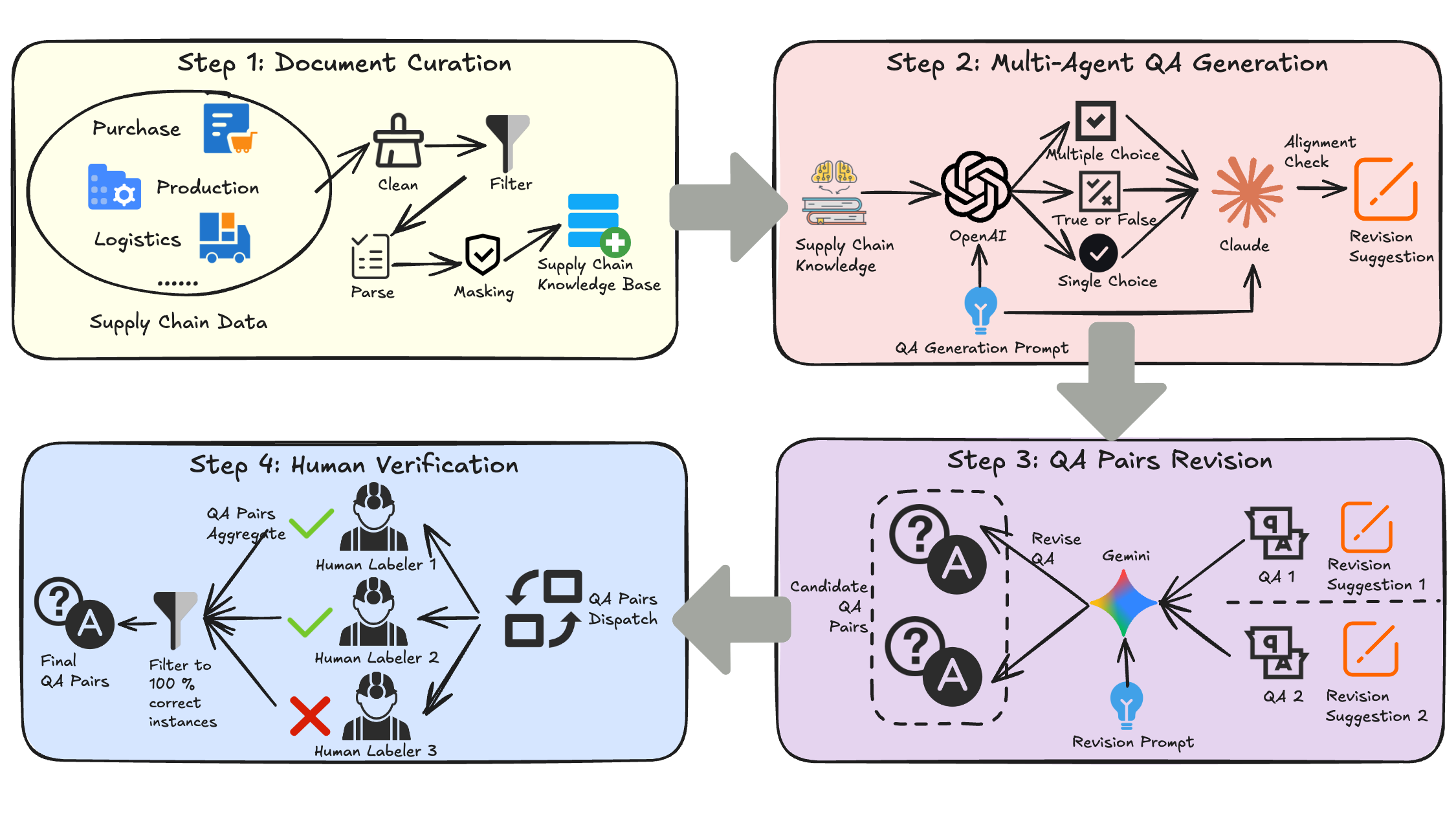}
    \caption{The dataset construction pipeline of SupChain-Bench follows a four-stage quality-assurance process. First, supply chain documents are curated and sanitized to construct a structured knowledge base. Second, a multi-agent LLM framework generates diverse QA candidates across multiple question formats. Third, these candidates undergo model-driven refinement to improve clarity, consistency, and factual correctness. Finally, human annotators perform strict verification, and only QA pairs that receive unanimous approval are included in the benchmark.}
    \label{fig:dataset-curation}
\end{figure*}

Recent research has increasingly shifted from general-purpose LLM performance metrics toward specialized benchmarks for supply chain and logistics applications, with an emphasis on domain-specific reasoning and decision-making capabilities.

One prominent line of work evaluates LLMs in specific operational roles. AIM-Bench assesses LLMs acting as inventory managers, revealing human-like biases such as the bullwhip effect and highlighting the potential of cognitive mitigation strategies \citep{Zhao2025AIMBench}. TransportBench evaluates models on transportation engineering tasks, exposing substantial variability in domain-specific reasoning performance \citep{syed2024benchmarkingcapabilitieslargelanguage}. For more complex sequential reasoning, EconLogicQA tests LLMs on multi-event economic scenarios, uncovering limitations in robust multi-step reasoning \citep{quan-liu-2024-econlogicqa}. Complementing these task-centric benchmarks, LalaEval introduces a human evaluation framework with domain-specific rubrics, providing standardized guidance for assessing logistics-oriented model behavior \citep{sun2024lalaevalholistichumanevaluation}.

Another stream of work focuses on demand forecasting and time-series reasoning. \citet{gruver2024largelanguagemodelszeroshot} demonstrate the zero-shot potential of LLMs for rapid forecasting. FoundTS \citep{li2025tsfmbenchcomprehensiveunifiedbenchmark} and MultiCast \citep{chatzigeorgakidis2024multicastzeroshotmultivariatetime} offer unified evaluation pipelines for multivariate time-series forecasting, while feature-centric benchmarks \citep{fons2024evaluatinglargelanguagemodels} examine how effectively LLMs capture trends and seasonality to improve interpretability and accuracy.

In the operations research (OR) domain, benchmarks primarily emphasize optimization and constraint satisfaction. ORQA evaluates LLM reasoning for industrial optimization challenges \citep{mostajabdaveh2025evaluatingllmreasoningoperations}, while R-ConstraintBench tests robustness under strict resource and time constraints \citep{jain2025rconstraintbenchevaluatingllmsnpcomplete}. Similarly, LLMs Can Schedule provides a supervised benchmark for job-shop scheduling problems \citep{abgaryan2024llmsschedule}. Beyond operational tasks, strategic decision-making benchmarks such as AIM-Bench (revisited for policy bias), BIBench, and InvestorBench extend evaluation to supplier assessment, business intelligence, and risk-sensitive decision making \citep{Zhao2025AIMBench, gupta-etal-2025-bi, li-etal-2025-investorbench}, while EconLogicQA further examines reasoning under cascading economic and supply chain events \citep{quan-liu-2024-econlogicqa}.

Despite these advances, existing benchmarks largely focus on isolated supply chain components, narrow operational roles, or generic decision-making tasks. In contrast, our benchmark provides an integrated, SCM-specific evaluation that jointly measures conceptual understanding and executable problem-solving. By combining structured Knowledge QA with realistic, long-horizon Tool Calling workflows, it captures both what a model knows and how effectively it can operationalize that knowledge in end-to-end supply chain processes. Moreover, evaluating tool use under both SOP-guided and unguided settings disentangles procedural compliance from autonomous planning ability, yielding a more comprehensive and practice-aligned assessment of LLM capabilities for supply chain management.

\section{Benchmark Design}
SupChain-Bench comprises two components: a QA Benchmark that evaluates domain understanding and reasoning over real-world supply-chain documents, and a Function-Calling Benchmark that measures a model’s ability to execute industrial SOPs via a simulated tool environment. Together, these components provide a holistic assessment of both knowledge comprehension and operational problem-solving capabilities.

\subsection{QA Dataset Curation} 
To ensure high-quality evaluation data, we developed a four-stage Human-in-the-Loop pipeline, summarized in Figure~\ref{fig:dataset-curation}, which progressively transforms raw operational documents into reliable QA pairs. The process begins with collecting and sanitizing expert-level documents to serve as a comprehensive knowledge base. A multi-model question generation stage then produces and iteratively refines QA pairs, leveraging successive models that generate, critique, and synthesize content. Candidate items undergo automated revision to ensure factual alignment, clarity, and appropriate complexity, and the final stage involves human domain experts performing a thorough review to guarantee correctness and linguistic quality. More detailed descriptions of each stage, as well as the rationale for our multi-model approach, can be found in Appendix~\ref{sec:qa_dataset_curation} and ~\ref{sec:rationle_for_generation}.

The resulting QA dataset is carefully balanced across three formats to evaluate different aspects of model comprehension. It includes 141 multiple-choice questions, 147 single-choice questions, and 147 true/false questions, providing a diverse set of challenges that reflect the complexity of real-world supply chain knowledge. This design ensures both the coverage of multiple functional domains and the ability to test nuanced understanding in a controlled evaluation setting.

\subsection{Function Calling Dataset Construction} 

To evaluate the ability of LLMs to invoke function calls in industrial settings, we also designed a Tool Calling benchmark that simulates a real-world supply chain database. We created a simulated environment containing core supply chain entities: trade order, fulfillment order, warehouse order, and cancellation context. This setup mimics a production database, allowing models to interact with data via API calls.

The questions in this benchmark are designed to mirror inquiries that require executing complex Standard Operating Procedures (SOPs). It is crucial to emphasize that these SOPs are workflows distilled by domain experts from extensive practical experience in resolving supply chain order issues. In the view of domain experts, strict adherence to these SOPs effectively guarantees the resolution of the corresponding problems. Consequently, if a model can accurately execute tools following these SOPs, it demonstrates the capability to solve real-world industrial challenges.

To generate these questions, we adapted the same multi-agent generation pipeline detailed in Section \ref{para:multi_agent_question_generation}. However, a key distinction lies in the input provided to the generation models.

Instead of curated documents, the process was as follows:
\begin{enumerate}[leftmargin=*, itemsep=0pt, labelsep=5pt, topsep=0pt]
    \item \textbf{Ground Truth Information Gathering:} For a given order, we first programmatically executed the corresponding SOP. This "oracle" script gathered all the raw information, such as order IDs, statuses, cancellation reasons, and error codes.
    \item \textbf{Question Synthesis:} This complete set of retrieved information was then fed as context into our multi-agent pipeline. The models' task was to synthesize a natural language question that would necessitate the execution of that specific SOP to be answered comprehensively.
\end{enumerate}

\begin{figure}[t]
    \centering
    \includegraphics[width=\linewidth]{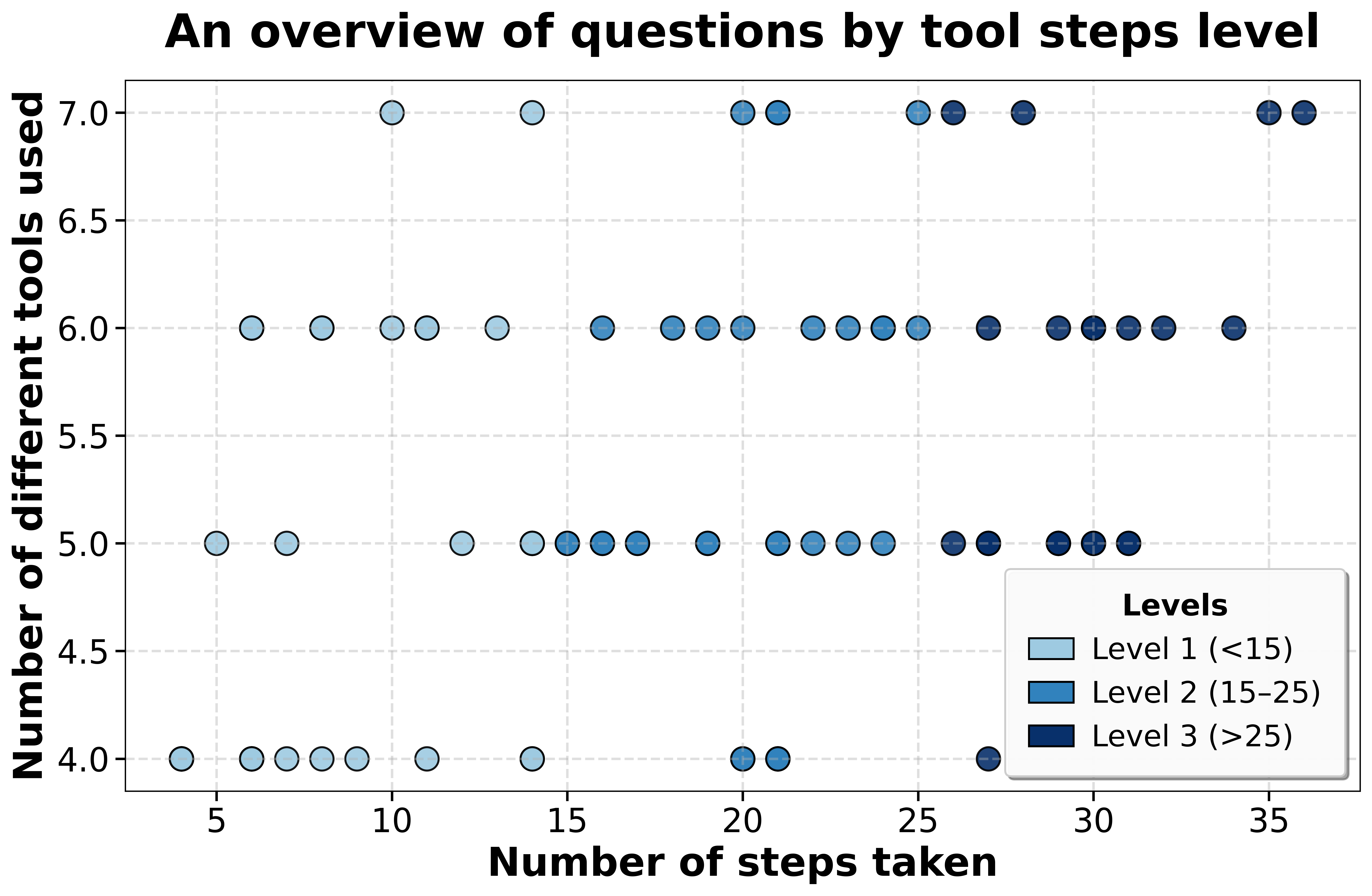}
    \caption{Distribution of function-calling questions by tool complexity in the dataset. Each point corresponds to a unique question, where the x-axis denotes the number of execution steps required and the y-axis indicates the number of distinct tools invoked.}
    \label{fig:tool_difficulty_distribution}
\end{figure}

Then, we produces questions spanning diverse complexity levels. As shown in Figure~\ref{fig:tool_difficulty_distribution}, the dataset exhibits three distinct tiers: Level 1 questions (<15 steps) cover basic information retrieval; Level 2 (15-25 steps) involve moderate multi-hop reasoning, such as cross-referencing order data with logistics events; and Level 3 (>25 steps) require complex orchestration of up to 7 distinct tools across 30+ execution steps. 

The evaluation then measures the \textbf{Information Retrieval Accuracy}. We compare the information retrieved by the model's tool calls against the Ground Truth set. This rigorously tests the model’s ability to handle multi-hop reasoning and long-chain tool execution. To achieve a high accuracy, a model must understand the dependencies between tools and strictly follow the industrial SOP to replicate the information gathered by our oracle script. Nevertheless, information retrieval accuracy alone does not fully reflect the model’s end-to-end performance, as it may fail to capture reasoning mistakes that arise even after successful tool execution. Therefore, in the appendix ~\ref{sec:additional_experiments}, we additionally report an LLM-as-a-judge-based evaluation of the final answer correctness.

\begin{table*}[t] 
\caption{Overall performance of different models on QA and tool-calling tasks. Results are reported under both context-free and context-augmented settings for QA, as well as SOP-free and SOP-guided settings for tool calling. Values shown in \textcolor{green!60!black}{green} indicate absolute accuracy gains when additional context or SOP guidance is provided, while values in \textcolor{red!80!black}{red} denote absolute accuracy drops under the same conditions.}
\centering
\small 
\setlength{\tabcolsep}{10pt} 
\begin{tabular}{l|cc|cc} 
\toprule
\multirow{2}{*}{\textbf{Model}} 
& \multicolumn{2}{c|}{\textbf{QA Accuracy (\%)}} 
& \multicolumn{2}{c}{\textbf{Tool Calling Accuracy (\%)}} \\ 
\cmidrule(lr){2-3} \cmidrule(lr){4-5}
& {w/o Context} & {w/ Context} 
& {w/o SOP} & {w/ SOP} \\ 
\hline
\textbf{\textit{Proprietary LLM}} & & & & \\
\quad GPT-4o             & 62.07 & 77.24 \gain{15.17} & 20.40 & 16.32 \loss{4.08} \\
\quad GPT-4.1-mini       & 68.05 & 80.46 \gain{12.41} & 11.22 & 28.57 \gain{17.35} \\
\quad GPT-4.1            & 68.97 & 82.76 \gain{13.79} & 12.24 & 16.32 \gain{4.08} \\
\quad GPT-O3-mini        & 65.06 & 80.23 \gain{15.17} & 9.18  & 14.28 \gain{5.10} \\
\quad GPT-5-nano         & 61.61 & 77.24 \gain{15.63} & 25.51 & 84.69 \gain{59.18} \\
\quad GPT-5-mini         & 66.90 & 83.91 \gain{17.01} & 46.93 & 86.73 \gain{39.80} \\
\quad GPT-5              & 66.21 & 83.91 \gain{17.70} & 35.71 & 84.69 \gain{48.98} \\
\quad Claude-3.5-Sonnet  & 67.81 & 80.45 \gain{12.64} & 11.22 & 29.59 \gain{18.37} \\
\quad Claude-3.7-Sonnet  & 62.30 & 78.16 \gain{15.86} & 35.71 & 40.09 \gain{4.38} \\
\quad Claude-4-Sonnet    & 66.43 & 82.07 \gain{15.64} & 31.63 & 39.79 \gain{8.16} \\
\quad Claude-4-Opus      & 66.70 & 80.69 \gain{13.99} & 26.53 & 34.69 \gain{8.16} \\
\quad Gemini-2.5-Flash   & 20.46 & 47.48 \gain{27.02} & 10.20 & 23.46 \gain{13.26} \\
\quad Gemini-2.5-Pro     & 66.67 & 80.69 \gain{14.02} & 11.22 & 37.56 \gain{26.34} \\
\midrule
\textbf{\textit{Open-Source LLM}} & & & & \\
\quad Deepseek-V3        & 67.75 & 79.77 \gain{12.02} & 10.20 & 33.63 \gain{23.43} \\
\quad Deepseek-R1        & 63.68 & 78.16 \gain{14.48} & 17.14  & 20.20 \gain{3.06} \\
\quad Kimi-K2-Instruct   & 64.83 & 80.68 \gain{15.85} & 12.24 & 10.20 \loss{2.04} \\
\quad Qwen3-Max          & 59.77 & 80.69 \gain{20.92} & 7.14  & 28.57 \gain{21.43} \\
\quad Qwen3-30B-A3B      & 56.55 & 77.93 \gain{21.38} & 5.10  & 8.16 \gain{3.06}  \\
\quad QwQ-32B            & 60.00 & 77.70 \gain{17.70} & 8.16  & 16.32 \gain{8.16} \\
\bottomrule
\end{tabular}
\label{tab:overall_llm}
\end{table*}

\section{Experiments}

\subsection{Settings}
We evaluate models in two settings: without external context and with context. In the without‑external‑context setting, we assess the model’s intrinsic domain knowledge. In the with‑context setting, we assess the model’s ability to process and reason over provided proprietary information.

For tasks based on knowledge documents, we use a unified prompt template, as shown in Table~\ref{tab:qa_prompt}. Both the system prompt and the user prompt explicitly specify the instructions given to the model. This template is applied consistently across single‑choice, multiple‑choice, and judgment (true/false) questions: the question stem and options are formatted in the same way, and the model is always instructed to output the final answer in a fixed pattern . After generation, we extract the model’s prediction from its output using regular expressions. More detailed examples of these prompts are provided in the Appendix.

When evaluating with background documents, for strict consistency, if a document exceeds 2,000 characters, we truncate it to the first 2,000 characters. The truncated document is prefixed to the user prompt. We verified that all task-relevant information (e.g., key facts required to answer questions) is contained within this window, ensuring that truncation does not introduce false negatives. This limit is set to simulate realistic input constraints in production environments.

For experiments involving tool usage, in the absence of a Standard Operating Procedure (SOP), the model only receives the user prompt and the schemas of the available tools. When an SOP is present, we additionally provide step‑by‑step instructions describing how to invoke the tools. Concrete examples for both settings are detailed in the Appendix.

\subsection{Main Results}
Our evaluation reveals two key insights regarding the behavior of modern LLMs in supply chain applications. First, the inclusion of domain-specific contextual information significantly enhances model performance across both question-answering and tool-calling tasks. Second, models differ substantially in their ability to persist through long sequences of tool executions, highlighting variations in long-horizon planning and adherence to procedural requirements. We elaborate on these findings in the following two sections.

\paragraph{Enhancing LLM with Supply Chain Context.}
In supply chain management, foundational domain knowledge is generally well captured by both open-source and proprietary LLMs, largely due to the abundance of publicly available data on common supply chain operations. As a result, performance on question-answering (QA) tasks without additional context is broadly comparable across different model architectures and training strategies. For example, GPT-4.1-mini and Claude-3.5-Sonnet achieve similar QA accuracies of 68.05\% and 67.81\%, respectively, indicating that diverse models possess comparable capabilities in understanding general supply chain knowledge.

Providing high-quality, real-world contextual information, however, leads to substantial performance improvements. GPT-5-mini, for instance, increases its QA accuracy from 66.90\% to 83.91\%, a gain of 17.01 percentage points, while other models exhibit improvements ranging from 12.02 to 21.38 percentage points. These results suggest that access to task-specific context consistently enhances model performance, regardless of the underlying architecture.

The impact of contextual information is even more pronounced in tool-calling tasks. When supplied with standard operating procedures, GPT-5-nano improves from 25.51\% to 84.69\%, and Qwen3-Max from 7.14\% to 28.57\%. Such dramatic gains highlight the critical role of realistic, high-quality procedural context in specialized applications, demonstrating that context-driven strategies can be more effective than model scaling alone.

Overall, these results indicate that in supply chain management, providing relevant contextual information is a practical and data-efficient strategy for achieving high model performance. High-quality supply chain management knowledge and realistic scenario data consistently yield substantial improvements. This emphasizes the importance of leveraging domain-specific information to build effective AI systems capable of handling complex, real-world tasks in professional environments.

\begin{figure*}[h]
    \centering
    \includegraphics[width=0.8\linewidth]{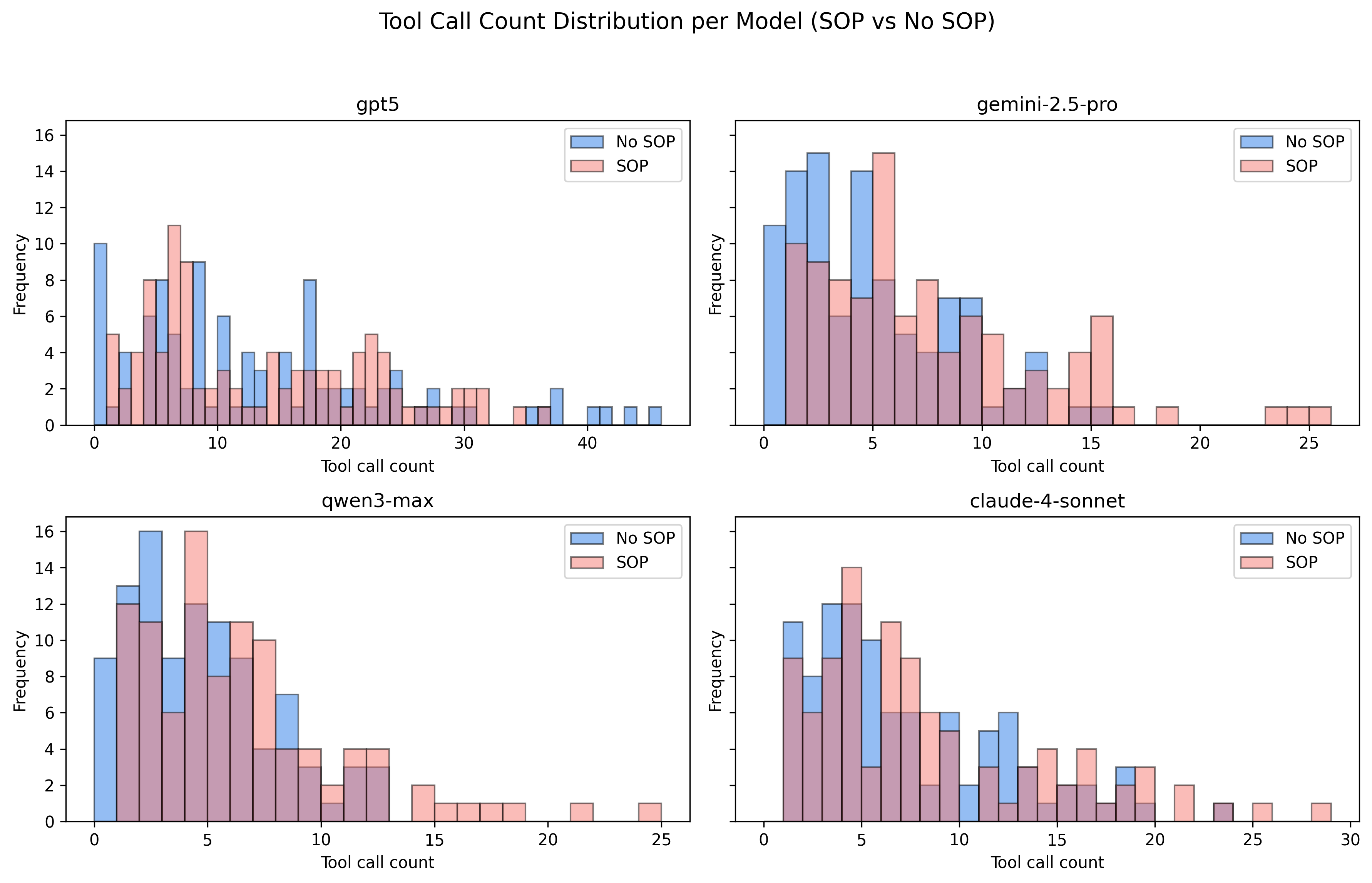}
    \caption{Overlaid histograms compare the number of tool calls per task for four models (gpt5, gemini-2.5-pro, qwen3-max, and claude-4-sonnet) under No SOP (blue) versus SOP (red). Across models, introducing an SOP generally shifts the distribution toward higher tool-call counts and produces heavier right tails, indicating longer and more persistent tool-use chains, while the no-SOP setting more often concentrates on shorter sequences and shows more early termination.}
    \label{fig:tool_distribution}
\end{figure*}

\paragraph{Long-Horizon Tool-Use Persistence Across LLMs.}
In function-calling tasks that require strict adherence to a predefined SOP, the LLM must execute long, sequential chains of tool calls—often exceeding twenty steps. Across these long-horizon settings, we observe substantial behavioral divergence between models. GPT models consistently follow the full instruction sequence, while Claude, Gemini, Qwen, and others frequently follow only part of the SOP and terminate early.

This indicates that, beyond short-horizon tool-use competence, current LLMs differ markedly in long-horizon compliance and execution persistence. Several models implicitly prioritize conversational efficiency or tool-use cost over the user’s explicit requirement for exhaustive coverage, thereby adopting a “good-enough stopping rule” even when it contradicts the prompt. 

As shown in Figure~\ref{tab:overall_llm}, GPT-5 exhibits strong goal retention and can execute long tool-call sequences even without explicit SOP guidance. Qualitative inspection reveals that its tool usage is neither redundant nor cyclical, indicating implicit state tracking and systematic traversal that closely mirrors SOP logic. This behavior suggests robust intrinsic planning that maintains alignment with global objectives while avoiding execution drift. In contrast, Claude-4-Sonnet, Gemini-2.5-Pro, and Qwen3-Max rely heavily on explicit procedural guidance. Although SOPs encourage longer tool-call sequences, these models still terminate prematurely relative to ground-truth workflows, reflecting a preference for conversational efficiency over task completion. Such behavior is consistent with metacognitive myopia or cost-biased action selection under long-horizon constraints.

Overall, in this multi-step diagnostic scenario involving repeated tool calls, the models exhibit clear long-horizon performance variance. GPT consistently executes complete tool chains even without SOPs, whereas other models often halt early despite explicit guidance. These differences indicate that the primary challenge in deploying LLM-based agents for real-world supply chain diagnostics lies not only in their reasoning ability but also in their reliability, endurance, and perseverance when executing extended, high-cost action sequences.

\subsection{Complexity-stratified Accuracy and Error Modes}
\begin{table}[t]
\caption{Tool calling accuracy across three difficulty levels with and without SOP guidance.}
\centering
\footnotesize        
\setlength{\tabcolsep}{4pt}   
\renewcommand{\arraystretch}{1} 

\begin{tabular}{l|ccc|ccc}
\toprule
\multirow{2}{*}{\textbf{Model}} 
& \multicolumn{3}{c|}{\textbf{w/o SOP (\%)}} 
& \multicolumn{3}{c}{\textbf{w/ SOP (\%)}} \\
\cmidrule(lr){2-4} \cmidrule(lr){5-7}
& \textbf{L1} & \textbf{L2} & \textbf{L3} & \textbf{L1} & \textbf{L2} & \textbf{L3} \\
\midrule
GPT-5            & 54.5 & 47.4 & 14.3 & 86.4 & 86.8 & 88.6 \\
Claude-4-Sonnet  & 31.8 & 36.8 & 28.6 & 45.5 & 50.0 & 28.6 \\
Gemini-2.5-Pro   & 18.2 & 5.3  & 14.3 & 45.5 & 47.4 & 25.7 \\
Deepseek-V3      & 18.2 & 5.3  & 11.4 & 50.0 & 55.3 & 11.4 \\
Kimi-K2-Instruct & 18.2 & 13.2 & 8.6  & 27.3 & 5.3  & 5.7  \\
Qwen3-Max        & 4.5  & 7.9  & 8.6  & 40.9 & 34.2 & 17.1 \\
\bottomrule
\end{tabular}
\label{tab:tool_calling_levels}
\end{table}

Table~\ref{tab:tool_calling_levels} shows the effect of SOP guidance on tool-calling accuracy across six state-of-the-art language models and three difficulty levels. Without SOP, all models perform poorly, especially on L2 and L3, often below 50\%. With SOP, accuracy improves substantially for most models: GPT-5 reaches 86.4\%, 86.8\%, and 88.6\% on L1–L3, while Deepseek-V3 and Claude-4-Sonnet also benefit notably on L1 and L2. Gemini-2.5-Pro and Kimi-K2-Instruct gain on simpler tasks but not on L3, highlighting large variability in how models exploit procedural guidance.

Error analysis shows that performance degradation with increasing complexity is driven by failures in multi-step reasoning, field-level fidelity, and handling of nested data. GPT-5 benefits from SOP at low complexity, but L2/L3 errors mainly arise from post-processing, where order status and cancellation metadata are dropped due to filtering or schema mismatches despite strong tool orchestration. Claude-4-Sonnet exhibits similar behavior, often normalizing anomalous states to valid enums and favoring schema conformity over factual accuracy. Qwen3-Max is highly sensitive to query focus, frequently omitting less salient fields, failing to enumerate multiple orders, and truncating long-text fields, reflecting a bias toward brevity and limited SOP utilization. Gemini-2.5-Pro shows weak schema alignment, with nested status and cancellation fields often lost unless explicitly requested.

Overall, accuracy declines from L1 to L3 for different reasons: GPT-5 and Claude suppress or over-normalize fields, Qwen loses track in multi-entity settings, and Gemini lacks strict schema governance. SOP mitigates some of these errors, but only for models that are already predisposed to follow structural constraints, underscoring that higher-level, parallel extraction tasks remain a key bottleneck for reliable tool-using agents. More detail error analysis could be found in Appendix~\ref{sec:error_analysis}.

\subsection{Efficiency Analysis}
Table ~\ref{tab:sop_comparison} reports both token consumption and accuracy for L1–L3, comparing runs with and without SOP. Across all models, SOP consistently improves accuracy, often turning low-performing settings into much more reliable ones, especially on L1 and L2. In terms of efficiency, the impact on token usage is not uniform: GPT‑5 becomes more token-efficient with SOP (fewer tokens at all levels), while several other models show a moderate increase in tokens when SOP is enabled. Overall, SOP primarily boosts effectiveness, and its efficiency trade-off depends on the specific model and task difficulty.

\begin{table}[ht]
\centering
\caption{Comparison of model performance with and without SOP guidance across three levels of tool complexity (L1--L3). For each model, we report the average number of generated tokens (Tok) and task accuracy (Acc.\%). Results highlight how SOP guidance affects both execution efficiency and success rates under increasing long-horizon complexity.}
\label{tab:sop_comparison}
\setlength{\tabcolsep}{2pt}          
\renewcommand{\arraystretch}{0.75}   
\footnotesize
\resizebox{\linewidth}{!}{%
\begin{tabular}{@{} l l ccccccc @{}}
\toprule
& & \multicolumn{3}{c}{\textbf{No SOP}} & \multicolumn{3}{c}{\textbf{With SOP}} \\
\cmidrule(lr){3-5} \cmidrule(l){6-8}
& Model & L1 & L2 & L3 & L1 & L2 & L3 \\
\midrule
\multirow{2}{*}{GPT-5}
    & Tok & 1123 & 1710 & 2202 & 871 & 1678 & 2126 \\
    & Acc & 54.5 & 47.4 & 14.3 & 86.4 & 86.8 & 88.6 \\
\midrule

\multirow{2}{*}{Claude-sonnet-4}
    & Tok & 888 & 1393 & 1826 & 1116 & 1678 & 1980 \\
    & Acc & 31.8 & 36.8 & 28.6 & 45.5 & 50.0 & 28.6 \\

\midrule

\multirow{2}{*}{Gemini-2.5-pro}
    & Tok & 520 & 1024 & 1214 & 720 & 1233 & 1282 \\
    & Acc & 18.2 & 5.3 & 14.3 & 45.5 & 47.4 & 25.7 \\

\midrule

\multirow{2}{*}{Kimi-K2-Instruct}
    & Tok & 857 & 1629 & 1988 & 1298 & 1571 & 1718 \\
    & Acc & 18.18  & 13.15  & 8.5 & 27.28 & 52.6  & 57.1 \\

\midrule

\multirow{2}{*}{Deepseek-v3}
    & Tok & 673 & 1059 & 1593 & 927 & 1285 & 1555 \\
    & Acc & 18.2 & 5.3 & 11.4 & 50.0 & 55.3 & 11.4 \\

\midrule

\multirow{2}{*}{Qwen3-max}
    & Tok & 518 & 991 & 1205 & 751 & 1135 & 1383 \\
    & Acc & 4.5 & 7.9 & 8.6 & 40.9 & 34.2 & 17.1 \\
\bottomrule
\end{tabular}%
}
\end{table}

\subsection{SupChain-ReAct}

In exploring long-horizon function-calling tasks, we observed a consistent pattern across models: when provided with a human-written SOP, tool-calling sequences tend to be longer, more stable, and more accurate (Figure~\ref{fig:tool_distribution}). This finding aligns with the intuition that SOPs offer explicit procedural scaffolding, effectively separating planning from execution; once a sufficiently clear plan is established, execution becomes more persistent and faithful. However, constructing such high-quality and robust SOPs requires substantial domain expertise and engineering effort, particularly in complex environments such as supply chain workflows. Notably, modern LLMs already demonstrate strong competence in function-calling semantics and tool-schema reasoning, suggesting that the primary remaining challenge lies not in tool understanding, but in procedural organization and planning.

These observations naturally lead to the following question: \textbf{\textit{If models already possess sufficient domain knowledge and tool-use competence, can they derive their own procedural guidance to support long-horizon execution?}}

Motivated by this question, we adopt \textbf{SupChain-ReAct}, a practical solution that enables models to synthesize procedural structure directly from their internal reasoning capabilities, without relying on externally authored SOPs. Rather than introducing new algorithmic components, SupChain-ReAct instantiates existing reasoning-and-acting paradigms in a manner that is particularly effective for our supply-chain-oriented setting.

Concretely, SupChain-ReAct builds upon the ReAct paradigm~\cite{yao2023reactsynergizingreasoningacting} and self-consistency-style sampling~\cite{wang2023selfconsistencyimproveschainthought} to improve planning stability and tool-calling persistence. Given a task prompt and a tool schema, the system executes multiple independent ReAct trajectories in parallel. Each trajectory alternates between reasoning steps and tool invocations under a shared prompt specification, and terminates upon producing a final answer or reaching a predefined step limit. In our implementation, we employ five parallel ReAct trajectories to balance computational efficiency and reasoning diversity. The final output is selected via a simple majority vote over the final textual answers produced by the successful trajectories. Notably, aggregation is performed only at the answer level, without requiring alignment or comparison of intermediate reasoning steps. This multi-path ReAct with voting setup introduces no external supervision or manually defined procedures, while empirically yielding more stable and accurate tool-calling behavior in our experiments.

\begin{table}[t]
\caption{Tool-calling performance of different models. Values shown in \textcolor{green!60!black}{green} denote absolute accuracy gains obtained from SOP guidance or the use of \textbf{SupChain-ReAct}, while values shown in \textcolor{red!80!black}{red} indicate absolute accuracy drops when SOP guidance is applied.}
\centering
\small
\setlength{\tabcolsep}{4pt}
\resizebox{\linewidth}{!}{%
\begin{tabular}{l|cc|g}
\toprule
\multirow{2}{*}{\textbf{Model}} 
& \multicolumn{3}{c}{\textbf{Tool Calling Accuracy (\%)}} \\
\cmidrule(lr){2-4}
& w/o SOP & w/ SOP & \cellcolor{white}SupChain-React \\
\midrule
GPT-4o             & 20.40 & 16.32 \loss{4.08}   & 40.81 \gain{20.41} \\
GPT-4.1-mini       & 11.22 & 28.57 \gain{17.35}  & 35.70 \gain{24.48} \\
GPT-4.1            & 12.24 & 16.32 \gain{4.08}   & 19.38 \gain{7.14}  \\
Claude-3.5-Sonnet  & 11.22 & 29.59 \gain{18.37}  & 40.81 \gain{29.59} \\
Claude-4-Sonnet    & 31.63 & 39.79 \gain{8.16}   & 75.51 \gain{43.88} \\
Claude-4-Opus      & 26.53 & 34.69 \gain{8.16}   & 75.51 \gain{48.98} \\
Gemini-2.5-Pro     & 11.22 & 37.56 \gain{26.34}  & 72.44 \gain{61.22} \\
Qwen3-Max          & 7.14  & 28.57 \gain{21.43}  & 33.67 \gain{26.53} \\
Qwen3-30B-A3B      & 5.10  & 8.16  \gain{3.06}   & 21.42 \gain{16.32} \\
\bottomrule
\end{tabular}%
}
\label{tab:supchain_react_tool_calling}
\end{table}

As shown in Table~\ref{tab:supchain_react_tool_calling}, SupChain-ReAct consistently achieves the highest tool-calling accuracy despite not relying on any handcrafted SOPs. These results indicate that, in our setting, enabling models to derive procedural guidance from their own reasoning processes is both effective and more generalizable than depending on externally authored procedures.

\section{Conclusion}
In this paper, we introduce SupChain-Bench, a realistic and principled benchmark for evaluating large language models in real-world supply chain management, providing a foundation for systematic study of their knowledge, reasoning, and operational capabilities.

\section*{Limitations}
Although SupChain-Bench evaluates LLM performance in realistic supply chain settings, it still has several limitations. First, it focuses on textual and tool-based interactions and does not incorporate multimodal signals such as sensor data, images, or real-time system logs commonly found in production supply chain systems. Second, while the benchmark is grounded in real-world workflows, practical deployments may involve additional sources of complexity, uncertainty, or coordination challenges that are not fully reflected in the current benchmark design.


\bibliography{references}

@misc{Zhao2025AIMBench,
      title={AIM-Bench: Evaluating Decision-making Biases of Agentic LLM as Inventory Manager}, 
      author={Xuhua Zhao and Yuxuan Xie and Caihua Chen and Yuxiang Sun},
      year={2025},
      eprint={2508.11416},
      archivePrefix={arXiv},
      primaryClass={cs.AI},
      url={https://arxiv.org/abs/2508.11416}, 
}

@article{Chen2025Optimized,
	article-number = {934},
	author = {Geng, Mingyang and Chen, Anping},
	doi = {10.3390/sym17060934},
	issn = {2073-8994},
	journal = {Symmetry},
	number = {6},
	title = {Optimized Scheduling for Multi-Drop Vehicle--Drone Collaboration with Delivery Constraints Using Large Language Models and Genetic Algorithms with Symmetry Principles},
	url = {https://www.mdpi.com/2073-8994/17/6/934},
	volume = {17},
	year = {2025}}

@misc{sun2024lalaevalholistichumanevaluation,
      title={LalaEval: A Holistic Human Evaluation Framework for Domain-Specific Large Language Models}, 
      author={Chongyan Sun and Ken Lin and Shiwei Wang and Hulong Wu and Chengfei Fu and Zhen Wang},
      year={2024},
      eprint={2408.13338},
      archivePrefix={arXiv},
      primaryClass={cs.HC},
      url={https://arxiv.org/abs/2408.13338}, 
}

@misc{syed2024benchmarkingcapabilitieslargelanguage,
      title={Benchmarking the Capabilities of Large Language Models in Transportation System Engineering: Accuracy, Consistency, and Reasoning Behaviors}, 
      author={Usman Syed and Ethan Light and Xingang Guo and Huan Zhang and Lianhui Qin and Yanfeng Ouyang and Bin Hu},
      year={2024},
      eprint={2408.08302},
      archivePrefix={arXiv},
      primaryClass={cs.AI},
      url={https://arxiv.org/abs/2408.08302}, 
}

@inproceedings{quan-liu-2024-econlogicqa,
    title = "{E}con{L}ogic{QA}: A Question-Answering Benchmark for Evaluating Large Language Models in Economic Sequential Reasoning",
    author = "Quan, Yinzhu  and
      Liu, Zefang",
    editor = "Al-Onaizan, Yaser  and
      Bansal, Mohit  and
      Chen, Yun-Nung",
    booktitle = "Findings of the Association for Computational Linguistics: EMNLP 2024",
    month = nov,
    year = "2024",
    address = "Miami, Florida, USA",
    publisher = "Association for Computational Linguistics",
    url = "https://aclanthology.org/2024.findings-emnlp.125/",
    doi = "10.18653/v1/2024.findings-emnlp.125",
    pages = "2273--2282"
}

@misc{gruver2024largelanguagemodelszeroshot,
      title={Large Language Models Are Zero-Shot Time Series Forecasters}, 
      author={Nate Gruver and Marc Finzi and Shikai Qiu and Andrew Gordon Wilson},
      year={2024},
      eprint={2310.07820},
      archivePrefix={arXiv},
      primaryClass={cs.LG},
      url={https://arxiv.org/abs/2310.07820}, 
}

@misc{li2025tsfmbenchcomprehensiveunifiedbenchmark,
      title={TSFM-Bench: A Comprehensive and Unified Benchmark of Foundation Models for Time Series Forecasting}, 
      author={Zhe Li and Xiangfei Qiu and Peng Chen and Yihang Wang and Hanyin Cheng and Yang Shu and Jilin Hu and Chenjuan Guo and Aoying Zhou and Christian S. Jensen and Bin Yang},
      year={2025},
      eprint={2410.11802},
      archivePrefix={arXiv},
      primaryClass={cs.LG},
      url={https://arxiv.org/abs/2410.11802}, 
}

@misc{chatzigeorgakidis2024multicastzeroshotmultivariatetime,
      title={MultiCast: Zero-Shot Multivariate Time Series Forecasting Using LLMs}, 
      author={Georgios Chatzigeorgakidis and Konstantinos Lentzos and Dimitrios Skoutas},
      year={2024},
      eprint={2405.14748},
      archivePrefix={arXiv},
      primaryClass={cs.LG},
      url={https://arxiv.org/abs/2405.14748}, 
}

@misc{fons2024evaluatinglargelanguagemodels,
      title={Evaluating Large Language Models on Time Series Feature Understanding: A Comprehensive Taxonomy and Benchmark}, 
      author={Elizabeth Fons and Rachneet Kaur and Soham Palande and Zhen Zeng and Tucker Balch and Manuela Veloso and Svitlana Vyetrenko},
      year={2024},
      eprint={2404.16563},
      archivePrefix={arXiv},
      primaryClass={cs.CL},
      url={https://arxiv.org/abs/2404.16563}, 
}

@misc{mostajabdaveh2025evaluatingllmreasoningoperations,
      title={Evaluating LLM Reasoning in the Operations Research Domain with ORQA}, 
      author={Mahdi Mostajabdaveh and Timothy T. Yu and Samarendra Chandan Bindu Dash and Rindranirina Ramamonjison and Jabo Serge Byusa and Giuseppe Carenini and Zirui Zhou and Yong Zhang},
      year={2025},
      eprint={2412.17874},
      archivePrefix={arXiv},
      primaryClass={cs.CL},
      url={https://arxiv.org/abs/2412.17874}, 
}

@misc{jain2025rconstraintbenchevaluatingllmsnpcomplete,
      title={R-ConstraintBench: Evaluating LLMs on NP-Complete Scheduling}, 
      author={Raj Jain and Marc Wetter},
      year={2025},
      eprint={2508.15204},
      archivePrefix={arXiv},
      primaryClass={cs.AI},
      url={https://arxiv.org/abs/2508.15204}, 
}

@misc{abgaryan2024llmsschedule,
      title={LLMs can Schedule}, 
      author={Henrik Abgaryan and Ararat Harutyunyan and Tristan Cazenave},
      year={2024},
      eprint={2408.06993},
      archivePrefix={arXiv},
      primaryClass={cs.AI},
      url={https://arxiv.org/abs/2408.06993}, 
}

@inproceedings{gupta-etal-2025-bi,
    title = "{BI}-Bench : A Comprehensive Benchmark Dataset and Unsupervised Evaluation for {BI} Systems",
    author = "Gupta, Ankush  and
      Aggarwal, Aniya  and
      Bithel, Shivangi  and
      Agarwal, Arvind",
    editor = "Rehm, Georg  and
      Li, Yunyao",
    booktitle = "Proceedings of the 63rd Annual Meeting of the Association for Computational Linguistics (Volume 6: Industry Track)",
    month = jul,
    year = "2025",
    address = "Vienna, Austria",
    publisher = "Association for Computational Linguistics",
    url = "https://aclanthology.org/2025.acl-industry.90/",
    doi = "10.18653/v1/2025.acl-industry.90",
    pages = "1287--1299",
    ISBN = "979-8-89176-288-6"
}

@inproceedings{li-etal-2025-investorbench,
    title = "{INVESTORBENCH}: A Benchmark for Financial Decision-Making Tasks with {LLM}-based Agent",
    author = "Li, Haohang  and
      Cao, Yupeng  and
      Yu, Yangyang  and
      Javaji, Shashidhar Reddy  and
      Deng, Zhiyang  and
      He, Yueru  and
      Jiang, Yuechen  and
      Zhu, Zining  and
      Subbalakshmi, K.p.  and
      Huang, Jimin  and
      Qian, Lingfei  and
      Peng, Xueqing  and
      Suchow, Jordan W.  and
      Xie, Qianqian",
    editor = "Che, Wanxiang  and
      Nabende, Joyce  and
      Shutova, Ekaterina  and
      Pilehvar, Mohammad Taher",
    booktitle = "Proceedings of the 63rd Annual Meeting of the Association for Computational Linguistics (Volume 1: Long Papers)",
    month = jul,
    year = "2025",
    address = "Vienna, Austria",
    publisher = "Association for Computational Linguistics",
    url = "https://aclanthology.org/2025.acl-long.126/",
    doi = "10.18653/v1/2025.acl-long.126",
    pages = "2509--2525",
    ISBN = "979-8-89176-251-0"
}

@misc{yao2023reactsynergizingreasoningacting,
      title={ReAct: Synergizing Reasoning and Acting in Language Models}, 
      author={Shunyu Yao and Jeffrey Zhao and Dian Yu and Nan Du and Izhak Shafran and Karthik Narasimhan and Yuan Cao},
      year={2023},
      eprint={2210.03629},
      archivePrefix={arXiv},
      primaryClass={cs.CL},
      url={https://arxiv.org/abs/2210.03629}, 
}

@misc{wang2023selfconsistencyimproveschainthought,
      title={Self-Consistency Improves Chain of Thought Reasoning in Language Models}, 
      author={Xuezhi Wang and Jason Wei and Dale Schuurmans and Quoc Le and Ed Chi and Sharan Narang and Aakanksha Chowdhery and Denny Zhou},
      year={2023},
      eprint={2203.11171},
      archivePrefix={arXiv},
      primaryClass={cs.CL},
      url={https://arxiv.org/abs/2203.11171}, 
}

@misc{li202512surveyreasoning,
      title={From System 1 to System 2: A Survey of Reasoning Large Language Models}, 
      author={Zhong-Zhi Li and Duzhen Zhang and Ming-Liang Zhang and Jiaxin Zhang and Zengyan Liu and Yuxuan Yao and Haotian Xu and Junhao Zheng and Pei-Jie Wang and Xiuyi Chen and Yingying Zhang and Fei Yin and Jiahua Dong and Zhiwei Li and Bao-Long Bi and Ling-Rui Mei and Junfeng Fang and Xiao Liang and Zhijiang Guo and Le Song and Cheng-Lin Liu},
      year={2025},
      eprint={2502.17419},
      archivePrefix={arXiv},
      primaryClass={cs.AI},
      url={https://arxiv.org/abs/2502.17419}, 
}

@inproceedings{wei-etal-2025-plangenllms,
    title = "{P}lan{G}en{LLM}s: A Modern Survey of {LLM} Planning Capabilities",
    author = "Wei, Hui  and
      Zhang, Zihao  and
      He, Shenghua  and
      Xia, Tian  and
      Pan, Shijia  and
      Liu, Fei",
    editor = "Che, Wanxiang  and
      Nabende, Joyce  and
      Shutova, Ekaterina  and
      Pilehvar, Mohammad Taher",
    booktitle = "Proceedings of the 63rd Annual Meeting of the Association for Computational Linguistics (Volume 1: Long Papers)",
    month = jul,
    year = "2025",
    address = "Vienna, Austria",
    publisher = "Association for Computational Linguistics",
    url = "https://aclanthology.org/2025.acl-long.958/",
    doi = "10.18653/v1/2025.acl-long.958",
    pages = "19497--19521",
    ISBN = "979-8-89176-251-0",
}

@misc{luo2025largelanguagemodelagent,
      title={Large Language Model Agent: A Survey on Methodology, Applications and Challenges}, 
      author={Junyu Luo and Weizhi Zhang and Ye Yuan and Yusheng Zhao and Junwei Yang and Yiyang Gu and Bohan Wu and Binqi Chen and Ziyue Qiao and Qingqing Long and Rongcheng Tu and Xiao Luo and Wei Ju and Zhiping Xiao and Yifan Wang and Meng Xiao and Chenwu Liu and Jingyang Yuan and Shichang Zhang and Yiqiao Jin and Fan Zhang and Xian Wu and Hanqing Zhao and Dacheng Tao and Philip S. Yu and Ming Zhang},
      year={2025},
      eprint={2503.21460},
      archivePrefix={arXiv},
      primaryClass={cs.CL},
      url={https://arxiv.org/abs/2503.21460}, 
}

@misc{openai_introducing_gpt5_2025_en,
  author       = {{OpenAI}},
  title        = {Introducing GPT-5},
  howpublished = {\url{https://openai.com/index/introducing-gpt-5/}},
  year         = {2025},
  month        = aug,
  note         = {Accessed: 2025-12-10},
}

@misc{yang2025qwen3technicalreport,
      title={Qwen3 Technical Report}, 
      author={An Yang and Anfeng Li and Baosong Yang and Beichen Zhang and Binyuan Hui and Bo Zheng and Bowen Yu and Chang Gao and Chengen Huang and Chenxu Lv and Chujie Zheng and Dayiheng Liu and Fan Zhou and Fei Huang and Feng Hu and Hao Ge and Haoran Wei and Huan Lin and Jialong Tang and Jian Yang and Jianhong Tu and Jianwei Zhang and Jianxin Yang and Jiaxi Yang and Jing Zhou and Jingren Zhou and Junyang Lin and Kai Dang and Keqin Bao and Kexin Yang and Le Yu and Lianghao Deng and Mei Li and Mingfeng Xue and Mingze Li and Pei Zhang and Peng Wang and Qin Zhu and Rui Men and Ruize Gao and Shixuan Liu and Shuang Luo and Tianhao Li and Tianyi Tang and Wenbiao Yin and Xingzhang Ren and Xinyu Wang and Xinyu Zhang and Xuancheng Ren and Yang Fan and Yang Su and Yichang Zhang and Yinger Zhang and Yu Wan and Yuqiong Liu and Zekun Wang and Zeyu Cui and Zhenru Zhang and Zhipeng Zhou and Zihan Qiu},
      year={2025},
      eprint={2505.09388},
      archivePrefix={arXiv},
      primaryClass={cs.CL},
      url={https://arxiv.org/abs/2505.09388}, 
}

@article{kumar2025inventory,
  author       = {Kumar, P. and Choubey, D.},
  title        = {Leveraging multivariate machine learning and large language models for multi-Billion Dollar inventory forecasting},
  journal      = {Global Journal of Engineering and Technology Advances},
  year         = {2025},
  volume       = {24},
  number       = {3},
  pages        = {034--042},
  doi          = {10.30574/gjeta.2025.24.3.0254},
  url          = {https://doi.org/10.30574/gjeta.2025.24.3.0254}
}

@misc{akbar2024smart,
  author       = {Akbar, D. and {\c{S}}im{\c{s}}ek, M.},
  title        = {Smart and Secure Supply Chains in the New Era Using Artificial Intelligence},
  year         = {2024},
  howpublished = {Preprint},
  doi          = {10.21203/rs.3.rs-4767072/v1},
  url          = {https://doi.org/10.21203/rs.3.rs-4767072/v1},
  note         = {Research Square}
}

@article{felder2025smart,
  author       = {Felder, M. and De Marchi, M. and Dallasega, P. and Rauch, E.},
  title        = {Smart Routing for Sustainable Supply Chain Networks: An AI and Knowledge Graph Driven Approach},
  journal      = {Applied Sciences},
  year         = {2025},
  volume       = {15},
  number       = {14},
  pages        = {8001},
  doi          = {10.3390/app15148001},
  url          = {https://doi.org/10.3390/app15148001}
}

\clearpage

\twocolumn[\DoToC]
\clearpage

\appendix

\section{Ethical Considerations}
This work aims to support the responsible study of large language models for supply chain management. The benchmark is constructed from real-world operational data that have been carefully anonymized and de-identified to remove sensitive or personally identifiable information. No private user data or confidential records are included.

The proposed benchmark is intended for research and evaluation purposes rather than direct deployment in operational systems. While LLMs show potential for assisting decision-making, their use in real-world supply chains may introduce risks related to reliability, data security, and regulatory compliance. To mitigate such concerns, our evaluation emphasizes controlled settings and highlights the importance of safeguards such as human oversight, auditing, and governance mechanisms.

By providing a transparent and standardized benchmark, this work aims to promote responsible research and informed deployment of LLM-based systems in operational contexts.

\section{Dataset Details}
\label{sec:dataset_details}

This data schema models a three-layer e-commerce supply chain: customer trade orders, fulfillment orders, and warehouse execution orders, with separate tables for errors and cancellation context plus a small set of tools that query and aggregate them. The tools expose a minimal API that lets an agent move from a front-end order ID to buyer identity, operational status, and detailed failure or cancellation reasons.

\paragraph{Core Table Relationships.}
The tables form a chain from the front-end order to physical operations: \texttt{TradeOrders} (customer orders) link to \texttt{FulfillmentOrders} (fulfillment jobs), which in turn link to \texttt{WarehouseOrders} (warehouse tasks) via foreign keys on \texttt{trade\_order\_id} and \texttt{fulfillment\_order\_id}. \texttt{ErrorLogs} attach error information either at warehouse level (keyed by the pair (\texttt{warehouse\_order\_id}, \texttt{fulfillment\_order\_id})) or at fulfillment level (keyed by \texttt{fulfillment\_order\_id}), while \texttt{CancellationContext} records who cancelled a fulfillment order and why, keyed by (\texttt{entity\_type} = \texttt{fulfillment\_order}, \texttt{entity\_id} = \texttt{fulfillment\_order\_id}). This separation between trade, fulfillment, and warehouse entities mirrors common e-commerce fulfillment models, where high-level orders are broken into operational jobs and then into location-specific warehouse tasks.

\paragraph{Supply Chain Scenarios Explain in Detail.}
\texttt{TradeOrders} represent what the buyer did on the front-end site: a \texttt{trade\_order\_id} is the external order number, and \texttt{buyer\_id} identifies the customer for downstream customer service, after-sales, and cancellation analysis. \texttt{FulfillmentOrders} represent how the platform decides to fulfill that trade order; the \texttt{fulfillment\_order\_id} is the main operational anchor, and \texttt{biz\_status} tracks whether fulfillment is still processing, has been shipped, cancelled, or delivered, matching typical fulfillment life cycles. Crucially, a single fulfillment order may be split into multiple physical shipments if items are sourced from different locations. \texttt{WarehouseOrders} represent concrete warehouse tasks such as picking, packing, dispatch, and delivery, with a fine-grained \texttt{status} that is mapped by \texttt{mapStatus()} into normalized business labels (for example \texttt{packing\_in\_progress}, \texttt{in\_transit}, \texttt{delivered}, or \texttt{error}) and an optional \texttt{error\_code} when something goes wrong (such as out-of-stock or undeliverable address).

\texttt{ErrorLogs} record the detailed cause of abnormal states: warehouse-level entries explain specific warehouse failures for a given pair (\texttt{warehouse\_order\_id}, \texttt{fulfillment\_order\_id}), while fulfillment-level entries capture problems that do not attach to a single warehouse order, such as routing failures or audit signals for fake shipping. \texttt{CancellationContext} captures who initiated a cancellation (for example buyer vs.\ seller), along with standardized reason codes and free-text explanations, so that support tools can attribute responsibility and analyze common cancellation patterns.

\paragraph{Tool Functions Over the Data.}
The tools expose the minimal path that an agent or system needs to answer key supply-chain questions without directly writing SQL over the tables. For traceability, \texttt{queryBuyerAndRelated()} starts from a trade-order ID and returns the \texttt{buyer\_id} plus the list of related fulfillment IDs and warehouse order IDs, effectively joining \texttt{TradeOrders}, \texttt{FulfillmentOrders}, and \texttt{WarehouseOrders}. For high-level fulfillment monitoring, \texttt{getFulfillmentStatus()} reads \texttt{FulfillmentOrders.biz\_status} and, when needed, aggregates the mapped \texttt{WarehouseOrders.status} values under the same \texttt{fulfillment\_order\_id} to decide whether the fulfillment is in progress, cancelled, abnormal, or completed.

For error analysis, \texttt{getWarehouseStatus()} returns the normalized warehouse status together with any raw \texttt{error\_code}, while \texttt{getWarehouseErrorDetails()} and \texttt{getErrorReason()} fetch the corresponding \{\texttt{code}, \texttt{text}\} explanations from \texttt{ErrorLogs} at warehouse and fulfillment levels. For cancellation understanding, \texttt{getCancelScenes()} reads the cancellation initiator (\texttt{cancel\_type}) from \texttt{CancellationContext}, and \texttt{getCancelErrorCode()} returns the structured and textual reasons (\texttt{reason\_code}, \texttt{reason\_text}). Finally, \texttt{checkFakeShipping()} examines fulfillment-level error codes and texts (for example entries like \texttt{FAKE\_SHIP} in \texttt{ErrorLogs}) to determine whether a given shipment should be flagged as suspected fake shipping.

\paragraph{Standard Operation Procedure}
To address user inquiries regarding a specific order, our system orchestrates a series of tools following a predefined Standard Operating Procedure (SOP). This process ensures that a comprehensive and accurate status is retrieved for any given \texttt{trade\_order\_id}. The workflow is as follows:

\begin{enumerate}[leftmargin=*, itemsep=0pt, labelsep=5pt, topsep=0pt]
    \item \textbf{Identifier Extraction:} The system first parses the user's query to extract the primary \texttt{trade\_order\_id}.

    \item \textbf{Order Information Retrieval:} Using the \texttt{trade\_order\_id}, the system queries our database to retrieve all associated order identifiers, including the \texttt{fulfillment\_id(s)} and \texttt{warehouse\_order\_id}. An order may be split into multiple fulfillments.

    \item \textbf{Status Check:} For each \texttt{fulfillment\_id} identified in the previous step, the system queries its real-time status.

    \item \textbf{Conditional Analysis:} The subsequent actions are contingent upon the status returned:
    \begin{itemize}[leftmargin=*, itemsep=0pt, labelsep=5pt, topsep=0pt]
        \item If the status is \texttt{cancelled}, the system proceeds to invoke two diagnostic tools: one to retrieve the cancellation reason (e.g., "customer initiated," "out of stock") and another to get the specific cancellation error code. An optional tool may also be triggered to check for signs of fraudulent shipping activity prior to cancellation.
        \item If the status is \texttt{error}, a dedicated tool is called to fetch the detailed error reason from the fulfillment system.
    \end{itemize}
\end{enumerate}

This template \ref{tab:sop_prompt} shows the detailed system prompt that will be used in every inference.

\section{Details of Dataset Construction}
\label{sec:qa_dataset_curation}
To ensure the evaluation captures real-world complexity and linguistic nuance, we constructed the dataset through a rigorous, multi-stage Human-in-the-Loop pipeline. As illustrated in Figure \ref{fig:dataset-curation}, the pipeline progressively transforms raw operational materials into high-quality QA pairs through document curation, multi-model question synthesis, and expert validation.

\paragraph{Step 1: Document Curation}
We first manually collected a comprehensive set of expert-level documents related to supply chain processes. All sensitive or proprietary information was meticulously filtered and anonymized to create a high-quality, sanitized document set that served as the knowledge base.The final corpus covers multiple functional domains across the end‑to‑end supply chain. It includes foundational concepts, key performance indicators, roles and entities, and detailed process for fulfillment and settlement, warehousing and consolidation operations, customs and cross‑border compliance, and first‑mile logistics. In addition, the collection contains routing and scheduling solutions.Together, these documents provide a rich, multi‑perspective description of real‑world supply chain workflows and supporting information systems.

\paragraph{Step 2: Multi-Model Question Generation}
\label{para:multi_agent_question_generation}
Using these documents and question as reference material, we developed a sophisticated, multi-agent pipeline to generate the final QA pair. This process, designed to mimic a collaborative expert workflow, involved three distinct stages:
\begin{enumerate}[leftmargin=*, itemsep=0pt, labelsep=5pt, topsep=0pt]
    \item \textbf{Initial Generation:} We prompted a capable model (\texttt{o3-mini}) to generate an initial question-answer pair based on the provided documents, prompt showed in Figure ~\ref{fig:qg-prompt}.
    \item \textbf{Adversarial Review and Refinement:} A second, advanced model (\texttt{Claude 4 Sonnet}) was tasked to act as a critic. It analyzed the generated QA pair and provided structured feedback for improvement, focusing on enhancing clarity, removing ambiguity, and increasing the cognitive complexity of the question, prompt showed in Figure ~\ref{fig:analysis-prompt}.
    \item \textbf{Final Synthesis:} A third, powerful model (\texttt{Gemini 2.5 Pro}) then synthesized the final, polished question by integrating the feedback from the critic model, prompt showed in Figure ~\ref{fig:final-prompt}.
\end{enumerate}
This multi-model approach proved highly effective, as the resulting questions were consistently rated by human reviewers as having greater nuance and quality compared to those from a single-pass generation process. 

\paragraph{Step 3: QA Pair Revision}
Before human review, all candidate QA pairs undergo an additional model-based revision stage. A dedicated model checks each item for factual alignment with the source documents and suggests targeted improvements, such as clarifying phrasing or adjusting distractors. Items that fail these automated checks are revised or discarded, resulting in a cleaner and more reliable pool for expert verification.

\paragraph{Step 4: Human Verification}
Finally, the filtered set of questions was reviewed by human domain experts. This last step served to eliminate any remaining awkward phrasing or technical inaccuracies, ensuring the final dataset's linguistic quality and correctness.


The resulting QA dataset is balanced across three formats to test different aspects of model comprehension:
\begin{itemize}[leftmargin=*, itemsep=0pt, labelsep=5pt, topsep=0pt]
    \item Multiple Choice: 141 questions
    \item Single Choice: 147 questions
    \item True/False: 147 questions
\end{itemize}

\subsection{Rationale for Multi-Model Question Generation Approach}
\label{sec:rationle_for_generation}
We adopted a heterogeneous multi-agent framework for question-answer generation after systematically evaluating different model configurations through expert assessment. Our methodology involved testing various agent compositions using 10 carefully curated documents, with domain experts ranking the quality of generated QA pairs across different experimental conditions.

Our empirical investigation revealed three key findings. First, when employing identical models across all three agent roles (e.g., exclusively using GPT o3-mini, Claude 4 Sonnet, or Gemini 2.5 Pro), the review and refinement stages frequently exhibited self-confirmation bias, accepting initial outputs that domain experts subsequently rated as lowest quality. This suggests that model homogeneity limits critical evaluation capacity within the pipeline.

Second, introducing a single heterogeneous model—regardless of its position in the three-stage pipeline—yielded modest quality improvements over homogeneous configurations. Notably, expert assessments showed no significant quality variation based on where the different model was positioned, indicating that diversity itself, rather than role-specific model assignment, drives improvement. This pattern persisted when introducing two heterogeneous models: positional placement remained inconsequential, while the presence of model diversity consistently enhanced output quality.

Third, maximizing model heterogeneity by deploying three distinct models produced the highest‑quality QA pairs according to expert evaluation. \textbf{We observed a clear monotonic improvement in quality as diversity increased, from homogeneous model combinations to one, then two, and finally three heterogeneous models.}

This improvement can be attributed to complementary model capabilities and reduced echo chamber effects. Different models possess distinct reasoning patterns, knowledge representations, and evaluation criteria. When heterogeneous models interact across generation, critique, and synthesis stages, they introduce diverse perspectives that challenge assumptions, identify blind spots, and surface alternative formulations. The adversarial review stage particularly benefits from this diversity, as critic models with different architectures can identify limitations that would be overlooked by architecturally identical reviewers. Furthermore, model heterogeneity mitigates the risk of systematic biases propagating through the pipeline, as each model's inherent limitations are compensated by others' strengths.

Given these findings, we selected o3-mini for initial generation, Claude 4 Sonnet for adversarial review, and Gemini 2.5 Pro for final synthesis—a configuration that maximizes model diversity while leveraging each model's specific capabilities within its assigned role.

\subsection{Prompt Overview}
\label{sec:prompt-overview}

\begin{table*}[t]
\centering
\small
\renewcommand\arraystretch{1.2} 

\begin{tabular}{p{0.95\textwidth}} 
    \toprule[1.5pt]
    \textbf{System Prompt: Supply Chain Diagnostic SOP} \\
    \midrule[1pt]
    
    \textbf{Role:} You are a deterministic Supply Chain Expert. Your job is to adhere to following standard operation procedure to answer user question. \\
    \vspace{0.3em} 
    \textbf{Deterministic Workflow:} \\
    
    \textbf{Step 1 --- Order-to-Fulfillment-and-Warehouse Expansion}
    \begin{itemize}[leftmargin=1.5em, nosep, before=\vspace{-0.3em}, after=\vspace{0.3em}]
        \item Call \texttt{query\_buyer\_and\_related} function to get related orders.
    \end{itemize}

    \textbf{Step 2 --- Fulfillment Status Resolution (per fulfillment\_id)}
    \begin{itemize}[leftmargin=1.5em, nosep, before=\vspace{-0.3em}, after=\vspace{0.3em}]
        \item If the user question does not specify any fulfillment order then, call \texttt{get\_fulfillment\_status} for each fulfillment order you get on first step.
    \end{itemize}

    \textbf{Step 3 --- Canceled Branch (If status == "canceled")}
    \begin{itemize}[leftmargin=1.5em, nosep, before=\vspace{-0.3em}, after=\vspace{0.3em}]
        \item Call \texttt{get\_cancel\_scenes(fulfillment\_id)} to get \texttt{cancelType}.
        \item Call \texttt{get\_cancel\_error\_code(fulfillment\_id)} to get \texttt{cancelErrorCode}/\texttt{cancelErrorMsg}.
        \item Optional: If \texttt{cancelType} is None/empty/unknown/other, call \texttt{check\_fake\_shipping}. Use only for internal analysis; do not add fields beyond the chosen Output Contract.
    \end{itemize}

    \textbf{Step 4 --- Error Branch (If status == "error")}
    \begin{itemize}[leftmargin=1.5em, nosep, before=\vspace{-0.3em}, after=\vspace{0.3em}]
        \item Call \texttt{get\_error\_reason(fulfillment\_id)} to get error code/text.
    \end{itemize}

    \textbf{Step 5 --- Warehouse-Level Enrichment} (for each known \texttt{warehouse\_order\_id} if the user question does not specify any)
    \begin{itemize}[leftmargin=1.5em, nosep, before=\vspace{-0.3em}, after=\vspace{0.3em}]
        \item Call \texttt{get\_warehouse\_status} and call \texttt{get\_warehouse\_error\_details} to check any error message.
        \item Prefer code/text from \texttt{get\_warehouse\_error\_details}; fallback to \texttt{get\_warehouse\_status.error} when detailed code is missing.
    \end{itemize}
    \\
    \bottomrule[1.5pt]
\end{tabular}%
\caption{The prompt template for the Standard Operating Procedure (SOP) used in our function calling task. The model is instructed to strictly follow the deterministic workflow involving sequential tool calls.}
\label{tab:sop_prompt}%
\end{table*}

\definecolor{bg_color}{RGB}{230, 242, 255} 
\sethlcolor{bg_color}
\begin{table}[t]
\centering
\footnotesize 
\renewcommand\arraystretch{1.2}

\begin{tabular}{p{0.15\linewidth} p{0.75\linewidth}}
    \toprule[1.5pt]
    \textbf{System} &
    {\ttfamily
    You are a helpful reasoning assistant. Let's think step by step. On the last line, output only the final answer in the format: '\#\#\#\# <LETTERS>'.
    } \\
    \midrule[1pt]
    \textbf{User} & 
    \colorbox{bg_color}{
        \parbox{0.95\linewidth}{ 
            \ttfamily \textbf{[Optional Context]} \\
            Background Information: \{truncated\_doc\}
        }
    } \vspace{0.5em} 
    
    {\ttfamily
    You will be given a multiple-choice question. First, think step by step \hl{based on the background information}. Finally, return ONLY the final answer on the last line in the format: \#\#\#\# <LETTERS> (e.g., AC).
    
    \vspace{0.3em}
    Question: \{question\}
    
    Options: \{options\}
    } \\
    \bottomrule[1.5pt]
\end{tabular}
\caption{Prompt template. The blue shaded area indicates the optional context (truncated to 2k chars) used only in the with-context setting.}
\label{tab:qa_prompt}
\end{table}

We employ primary prompt templates in our study—multi-model question generation, and multiple-choice QA—together with a deterministic Standard Operating Procedure (SOP) prompt for the tool-calling setting. Tables introduced in this section provide the exact templates used in our experiments.

\paragraph{Multiple-Choice QA Prompt.}
QA prompt: Table \ref{tab:qa_prompt} contains the template used for the multiple-choice evaluation stage. The prompt instructs the model to engage in step-by-step reasoning while committing to output only the final answer in a fixed format. When background context is provided, the template explicitly emphasizes grounding the reasoning in the given context. This setup minimizes stylistic variance and standardizes the evaluation interface across models.

\paragraph{SOP Prompt for Tool-Calling.}
SOP prompt: Table \ref{tab:sop_prompt} provides the deterministic SOP prompt used in our function-calling experiments. Unlike the free-form reasoning templates above, this prompt enforces a fixed operational workflow, requiring the model to perform a series of tool calls in a predefined order. Each step of the SOP corresponds to an operational requirement in the supply-chain diagnostic pipeline, including order expansion, fulfillment-state resolution, cancellation and error investigation, and warehouse-level enrichment. By constraining the model to follow this strictly ordered procedure, the SOP prompt enables reliable, reproducible evaluation in settings where correctness depends on precise tool invocation rather than open-ended natural-language reasoning.

\subsection{Error Analysis}
\label{sec:error_analysis}
Our error analysis shows that the dominant failure mode is missing warehouse status (4,215 cases). This is primarily driven by \textbf{schema mismatches}, showed in table ~\ref{tab:sop_prompt_t1002_full_trace} (the same concept appears under different keys such as status instead of warehouse status, \textbf{multi-entity extraction gaps}, showed in table ~\ref{tab:multi_entity_extraction_gaps} (per-fulfillment arrays of multiple work orders were not iterated, or aggregation/merge logic overwrote WO-level fields).The second-largest category, tool usage issues (974), reflects \textbf{faithfulness error}, showed in table~\ref{tab:sop_prompt_temporal_inconsistency} (model response is not consistent with tool response). Cancellation-related fields were also frequently missing: buyer cancel reason (525) was lost due to free-text truncation/sanitization, FO→WO propagation/join gaps, conditional gating based on question wording or cancel scene, and key-variant normalization failures; cancel scene (395) was omitted for similar FO-level linkage issues, enum/alias mismatches, and conservative conflict handling that defaulted to null. Smaller residual categories include general extraction failures (280) from missed nested selectors, non-iterated lists, tool errors, and merge/projection mistakes; normalization issues (150) from incomplete ontologies and strict whitelists without fallback raw-status retention; status mismatches (35) where fraud/anomaly flags were conflated with lifecycle stages due to precedence heuristics;

\subsection{License}
The code of this paper is released under the Apache License, Version 2.0.Apache-2.0 permits use, reproduction, modification, and distribution, and it includes an express patent license from contributors, subject to the license conditions (e.g., retaining copyright/license notices and marking significant changes). The accompanying dataset (and any documentation, unless otherwise stated) is also made available under Apache-2.0.

\begin{table*}[t]
\centering
\scriptsize           
\renewcommand\arraystretch{1.0}  

\begin{tabular}{p{0.95\linewidth}}
\toprule[1.5pt]
\textbf{System Prompt: Supply Chain Diagnostic SOP} \\
\midrule[1pt]

\textbf{Deterministic Workflow (summary):} The model follows a multi-step pipeline that expands trade orders to fulfillment and warehouse orders, resolves fulfillment status, handles canceled and error branches, and enriches warehouse-level information. \\
\midrule

\textbf{User Question:} Given that trade order \texttt{T1002} is associated with multiple fulfillment orders, each showing a business status of \texttt{in\_transit} despite containing warehouse orders in disparate states (including one already \texttt{delivered} and others still \texttt{packing\_in\_progress}), what does this indicate about the granularity of status tracking in multi-warehouse fulfillment scenarios and how should the overall fulfillment progress be interpreted? \\

\textbf{Model Output (summary):} The model finds that \texttt{FO2003} and \texttt{FO2004} remain \texttt{in\_transit} while other fulfillments are cancelled, and concludes that business status is tracked per fulfillment order, which may not strictly reflect heterogeneous warehouse-order states but still signals overall “in transit” progress. \\

\textbf{Tool Execution Trace:} \\
\texttt{[} \\
\hspace*{1em}\texttt{\{} \\
\hspace*{2em}\texttt{"step": 1,} \\
\hspace*{2em}\texttt{"name": "query\_buyer\_and\_related",} \\
\hspace*{2em}\texttt{"arguments": \{"order\_id": "T1002"\},} \\
\hspace*{2em}\texttt{"output": \{} \\
\hspace*{3em}\texttt{"buyer\_id": \{"id": 90001\},} \\
\hspace*{3em}\texttt{"related\_item": [} \\
\hspace*{4em}\texttt{\{"fulfillment\_id": "FO2002", "warehouse\_order\_id": "WO3002"\},} \\
\hspace*{4em}\texttt{\{"fulfillment\_id": "FO2003", "warehouse\_order\_id": "WO3003"\},} \\
\hspace*{4em}\texttt{\{"fulfillment\_id": "FO2003", "warehouse\_order\_id": "WO3004"\},} \\
\hspace*{4em}\texttt{\{"fulfillment\_id": "FO2003", "warehouse\_order\_id": "WO3005"\},} \\
\hspace*{4em}\texttt{\{"fulfillment\_id": "FO2004", "warehouse\_order\_id": "WO3006"\},} \\
\hspace*{4em}\texttt{\{"fulfillment\_id": "FO2004", "warehouse\_order\_id": "WO3007"\},} \\
\hspace*{4em}\texttt{\{"fulfillment\_id": "FO2004", "warehouse\_order\_id": "WO3008"\},} \\
\hspace*{4em}\texttt{\{"fulfillment\_id": "FO2005", "warehouse\_order\_id": "WO3009"\},} \\
\hspace*{4em}\texttt{\{"fulfillment\_id": "FO2006", "warehouse\_order\_id": "WO3010"\},} \\
\hspace*{4em}\texttt{\{"fulfillment\_id": "FO2006", "warehouse\_order\_id": "WO3011"\},} \\
\hspace*{4em}\texttt{\{"fulfillment\_id": "FO2006", "warehouse\_order\_id": "WO3012"\}} \\
\hspace*{3em}\texttt{]} \\
\hspace*{2em}\texttt{\}} \\
\hspace*{1em}\texttt{\},} \\
\hspace*{1em}\texttt{\{} \\
\hspace*{2em}\texttt{"step": 2,} \\
\hspace*{2em}\texttt{"name": "get\_fulfillment\_status",} \\
\hspace*{2em}\texttt{"arguments": \{"fulfillment\_id": "FO2002"\},} \\
\hspace*{2em}\texttt{"output": \{"status": "cancelled"\}} \\
\hspace*{1em}\texttt{\},} \\
\hspace*{1em}\texttt{\{} \\
\hspace*{2em}\texttt{"step": 3,} \\
\hspace*{2em}\texttt{"name": "get\_fulfillment\_status",} \\
\hspace*{2em}\texttt{"arguments": \{"fulfillment\_id": "FO2003"\},} \\
\hspace*{2em}\texttt{"output": \{"status": "in\_transit"\}} \\
\hspace*{1em}\texttt{\},} \\
\hspace*{1em}\texttt{\{} \\
\hspace*{2em}\texttt{"step": 4,} \\
\hspace*{2em}\texttt{"name": "get\_fulfillment\_status",} \\
\hspace*{2em}\texttt{"arguments": \{"fulfillment\_id": "FO2004"\},} \\
\hspace*{2em}\texttt{"output": \{"status": "in\_transit"\}} \\
\hspace*{1em}\texttt{\},} \\
\hspace*{1em}\texttt{\{} \\
\hspace*{2em}\texttt{"step": 5,} \\
\hspace*{2em}\texttt{"name": "get\_cancel\_scenes",} \\
\hspace*{2em}\texttt{"arguments": \{"fulfillment\_id": "FO2003"\},} \\
\hspace*{2em}\texttt{"output": \{"cancelType": null\}} \\
\hspace*{1em}\texttt{\},} \\
\hspace*{1em}\texttt{\{} \\
\hspace*{2em}\texttt{"step": 6,} \\
\hspace*{2em}\texttt{"name": "get\_fulfillment\_status",} \\
\hspace*{2em}\texttt{"arguments": \{"fulfillment\_id": "FO2005"\},} \\
\hspace*{2em}\texttt{"output": \{"status": "cancelled"\}} \\
\hspace*{1em}\texttt{\},} \\
\hspace*{1em}\texttt{\{} \\
\hspace*{2em}\texttt{"step": 7,} \\
\hspace*{2em}\texttt{"name": "get\_fulfillment\_status",} \\
\hspace*{2em}\texttt{"arguments": \{"fulfillment\_id": "FO2006"\},} \\
\hspace*{2em}\texttt{"output": \{"status": "cancelled"\}} \\
\hspace*{1em}\texttt{\},} \\
\hspace*{1em}\texttt{\{} \\
\hspace*{2em}\texttt{"step": 8,} \\
\hspace*{2em}\texttt{"name": "get\_cancel\_scenes",} \\
\hspace*{2em}\texttt{"arguments": \{"fulfillment": "FO2004"\},} \\
\hspace*{2em}\texttt{"output": \{"error": "fufillment is not a right input argument"\}} \\
\hspace*{1em}\texttt{\}} \\
\texttt{]} \\
\bottomrule[1.5pt]
\end{tabular}%
\caption{Schema-mismatch example for trade order T1002: the final tool call passes \texttt{fulfillment} instead of the required \texttt{fulfillment id}, triggering a schema validation error. Rather than correcting the argument and retrying the function call, the model terminates the function-calling sequence prematurely, and its final response also fails to disclose that no relevant information was retrieved during execution.}
\label{tab:sop_prompt_t1002_full_trace}
\end{table*}

\begin{table*}[t]
\centering
\small
\renewcommand\arraystretch{1.0}

\begin{tabular}{p{0.95\linewidth}}
\toprule[1.5pt]
\textbf{System Prompt: Supply Chain Diagnostic SOP} \\
\midrule[1pt]

\textbf{Deterministic Workflow (summary):} Same as the SOP shown in the previous table. \\
\midrule

\textbf{User Question:} Given that fulfillment order \texttt{FO2055} for trade order \texttt{T1018} is marked as \texttt{canceled}, why do its associated warehouse orders show statuses of \texttt{in\_transit} and \texttt{delivered}, indicating that physical fulfillment had already progressed or completed prior to cancellation? \\

\textbf{Model Output (compressed):} \texttt{FO2055}(\texttt{T1018}) links $\texttt{WO3108}=\texttt{in\_transit}$, $\texttt{WO3109}=\texttt{delivered}$, $\texttt{WO3110}=\texttt{delivered}$; cancellation $\texttt{cancelType}=\texttt{SELLER}$ with message ``Due to a last-minute change of plans, the buyer decided that the item was no longer necessary and canceled the order.'' This suggests cancellation was recorded after physical operations had started/finished and updates can be asynchronous. \\

\textbf{Tool Execution Trace:} \\
\begin{lstlisting}[basicstyle=\ttfamily\fontsize{6}{7}\selectfont, breaklines=true, breakatwhitespace=true]
[
  {
    "step": 1,
    "name": "query_buyer_and_related",
    "arguments": {"order_id": "T1018"},
    "output": {
      "buyer_id": {"id": 90017},
      "related_item": [
        {"fulfillment_id": "FO2055", "warehouse_order_id": "WO3108"},
        {"fulfillment_id": "FO2055", "warehouse_order_id": "WO3109"},
        {"fulfillment_id": "FO2055", "warehouse_order_id": "WO3110"}
      ]
    }
  },
  {
    "step": 2,
    "name": "get_fulfillment_status",
    "arguments": {"fulfillment_id": "FO2055"},
    "output": {"status": "cancelled"}
  },
  {
    "step": 3,
    "name": "get_cancel_scenes",
    "arguments": {"fulfillment_id": "FO2055"},
    "output": {"cancelType": "SELLER"}
  },
  {
    "step": 4,
    "name": "get_cancel_error_code",
    "arguments": {"fulfillment_id": "FO2055"},
    "output": {
      "cancelErrorCode": null,
      "cancelErrorMsg": "Due to a last-minute change of plans, the buyer decided that the item was no longer necessary"
    }
  },
  {
    "step": 5,
    "name": "get_warehouse_status",
    "arguments": {"fulfillment_id": "FO2055", "warehouse_order_id": "WO3108"},
    "output": {"status": "in_transit", "error": null}
  },
  {
    "step": 6,
    "name": "get_warehouse_error_details",
    "arguments": {"fulfillment_id": "FO2055", "warehouse_order_id": "WO3108"},
    "output": {"code": null, "text": null}
  },
  {
    "step": 7,
    "name": "get_warehouse_status",
    "arguments": {"fulfillment_id": "FO2055", "warehouse_order_id": "WO3109"},
    "output": {"status": "delivered", "error": null}
  },
  {
    "step": 8,
    "name": "get_warehouse_error_details",
    "arguments": {"fulfillment_id": "FO2055", "warehouse_order_id": "WO3109"},
    "output": {"code": null, "text": null}
  }
]
\end{lstlisting} \\
\bottomrule[1.5pt]
\end{tabular}%
\caption{Multi-Entity Extraction Gaps: After step 8, the model should query \texttt{WO3110} for its status and error details, but the function-calling sequence terminates at this point without issuing the required tool call. The final user-facing response also omits any mention of \texttt{WO3110}, failing to disclose that the relevant information was never retrieved.}
\label{tab:multi_entity_extraction_gaps}
\end{table*}

\begin{table*}[t]
\centering
\small
\renewcommand\arraystretch{1.0}

\begin{tabular}{p{0.95\linewidth}}
\toprule[1.5pt]
\textbf{System Prompt: Supply Chain Diagnostic SOP} \\
\midrule[1pt]

\textbf{Deterministic Workflow (summary):} Same as the SOP shown in the previous table. \\
\midrule

\textbf{User Question:} Given that fulfillment order \texttt{FO2055} for trade order \texttt{T1018} is marked as \texttt{canceled}, why do its associated warehouse orders show statuses of \texttt{in\_transit} and \texttt{delivered}, indicating that physical fulfillment had already progressed or completed prior to cancellation? \\

\textbf{Model Output:} \texttt{FO2055}(\texttt{T1018}) links $\texttt{WO3108}=\texttt{cancelled}$, $\texttt{WO3109}=\texttt{delivered}$, $\texttt{WO3110}=\texttt{delivered}$; cancellation $\texttt{cancelType}=\texttt{SELLER}$ with message ``Due to a last-minute change of plans, the buyer decided that the item was no longer necessary and canceled the order.'' This suggests cancellation was recorded after physical operations had started/finished and updates can be asynchronous. \\

\textbf{Tool Execution Trace:} \\
\begin{lstlisting}[basicstyle=\ttfamily\fontsize{6}{7}\selectfont, breaklines=true, breakatwhitespace=true]
[
  {
    "step": 1,
    "name": "query_buyer_and_related",
    "arguments": {"order_id": "T1018"},
    "output": {
      "buyer_id": {"id": 90017},
      "related_item": [
        {"fulfillment_id": "FO2055", "warehouse_order_id": "WO3108"},
        {"fulfillment_id": "FO2055", "warehouse_order_id": "WO3109"},
        {"fulfillment_id": "FO2055", "warehouse_order_id": "WO3110"}
      ]
    }
  },
  {
    "step": 2,
    "name": "get_fulfillment_status",
    "arguments": {"fulfillment_id": "FO2055"},
    "output": {"status": "cancelled"}
  },
  {
    "step": 3,
    "name": "get_cancel_scenes",
    "arguments": {"fulfillment_id": "FO2055"},
    "output": {"cancelType": "SELLER"}
  },
  {
    "step": 4,
    "name": "get_cancel_error_code",
    "arguments": {"fulfillment_id": "FO2055"},
    "output": {
      "cancelErrorCode": null,
      "cancelErrorMsg": "Due to a last-minute change of plans, the buyer decided that the item was no longer necessary and cancelled the order."
    }
  },
  {
    "step": 5,
    "name": "get_warehouse_status",
    "arguments": {"fulfillment_id": "FO2055", "warehouse_order_id": "WO3108"},
    "output": {"status": "in_transit", "error": null}
  },
  {
    "step": 6,
    "name": "get_warehouse_error_details",
    "arguments": {"fulfillment_id": "FO2055", "warehouse_order_id": "WO3108"},
    "output": {"code": null, "text": null}
  },
  {
    "step": 7,
    "name": "get_warehouse_status",
    "arguments": {"fulfillment_id": "FO2055", "warehouse_order_id": "WO3109"},
    "output": {"status": "delivered", "error": null}
  },
  {
    "step": 8,
    "name": "get_warehouse_error_details",
    "arguments": {"fulfillment_id": "FO2055", "warehouse_order_id": "WO3109"},
    "output": {"code": null, "text": null}
  }
]
\end{lstlisting}
\\
\bottomrule[1.5pt]
\end{tabular}%
\caption{Although the tool correctly reports \texttt{WO3108} as \texttt{"in transit"} at query time, the model later changes the status in its final response and incorrectly states that \texttt{WO3108} is \texttt{"cancelled"}.}
\label{tab:sop_prompt_temporal_inconsistency}
\end{table*}

\begin{table*}
\caption{Tool calling accuracy across three difficulty levels (Level 1, Level 2, Level 3) with and without SOP guidance.}
\centering
\small 
\setlength{\tabcolsep}{8pt}
\begin{tabular}{l|ccc|ccc}
\toprule
\multirow{2}{*}{\textbf{Model}} 
& \multicolumn{3}{c|}{\textbf{Tool Calling Accuracy w/o SOP (\%)}} 
& \multicolumn{3}{c}{\textbf{Tool Calling Accuracy w/ SOP (\%)}} \\ 
\cmidrule(lr){2-4} \cmidrule(lr){5-7}
& {Level 1} & {Level 2} & {Level 3}
& {Level 1} & {Level 2} & {Level 3} \\ 
\hline
GPT-4.1            & 22.7 & 5.3  & 11.4 & 40.9 & 10.5 & 8.6 \\
GPT-4.1-mini        & 18.2 & 10.5 & 11.4 & 45.5 & 31.6 & 17.1 \\
GPT-4o             & 18.2 & 31.6 & 11.4 & 31.8 & 10.5 & 14.3 \\
GPT-O3-mini           & 22.7 & 5.3  & 5.7  & 36.4 & 10.5 & 5.7 \\
GPT-5              & 54.5 & 47.4  & 14.3  & 86.4 & 86.8 & 88.6 \\
GPT-5-mini         & 63.6 & 60.5 & 25.7 & 81.8 & 94.7 & 88.6 \\
GPT-5-nano          & 22.7 & 31.6 & 22.9 & 81.8 & 89.5 & 88.6 \\
Claude-Sonnet-3.5    & 18.2 & 7.9  & 11.4 & 54.5 & 23.7 & 22.9 \\
Claude-Sonnet-3.7    & 54.5 & 34.2 & 28.6 & 59.1 & 42.1 & 31.4 \\
Claude-Sonnet-4    & 31.8 & 36.8 & 28.6 & 45.5 & 50.0 & 28.6 \\
Claude-Opus-4     & 31.8 & 39.5 & 11.4 & 59.1 & 42.1 & 14.3 \\
Gemini-2.5-flash               & 13.6 & 5.3  & 14.3 & 27.3 & 28.9 & 17.1 \\
Gemini-2.5-pro                 & 18.2 & 5.3 & 14.3  & 45.5 & 47.4 & 25.7 \\
Deepseek-V3        & 18.2 & 5.3 & 11.4 & 50 & 55.3 & 11.4 \\
Qwen3-max                      & 4.5  & 7.9  & 8.6  & 40.9 & 34.2 & 17.1 \\
Qwen3-30B-A3B                  & 4.5  & 0.0  & 11.4 & 13.6 & 5.3  & 8.6 \\
Kimi-K2-Instruct                  & 18.18  & 13.15  & 8.5 & 27.28 & 52.6  & 57.1 \\
\bottomrule
\end{tabular}
\label{tab:tool_calling_levels_big_tables}
\end{table*}

\begin{table*}[htbp]
\centering
\caption{GPT-series Average Token Length by SOP status, correctness, and level}
\label{tab:gpt_avg_combined_sop}
\setlength{\tabcolsep}{3pt}
\resizebox{\textwidth}{!}{%
\begin{tabular}{lrrrrrrrrrrrr}
\hline
 & \multicolumn{4}{c}{Level 1} & \multicolumn{4}{c}{Level 2} & \multicolumn{4}{c}{Level 3} \\
Model & \multicolumn{2}{c}{No-SOP} & \multicolumn{2}{c}{SOP} & \multicolumn{2}{c}{No-SOP} & \multicolumn{2}{c}{SOP} & \multicolumn{2}{c}{No-SOP} & \multicolumn{2}{c}{SOP} \\
 & Correct & Incorrect & Correct & Incorrect & Correct & Incorrect & Correct & Incorrect & Correct & Incorrect & Correct & Incorrect \\
\hline
O3-mini & 409.67 & 258.11 & 384.50 & 310.38 & 507.29 & 438.18 & 401.00 & 478.67 & 351.00 & 686.62 & 487.00 & 637.56 \\
GPT-4.1 & 535.00 & 264.33 & 504.25 & 281.25 & 656.00 & 405.20 & 536.00 & 472.36 & 756.00 & 582.50 & 666.00 & 724.88 \\
GPT-4.1-mini & 456.00 & 251.78 & 573.00 & 275.88 & 719.00 & 411.10 & 577.25 & 470.22 & 815.50 & 611.12 & 793.50 & 663.00 \\
GPT-4o & 282.50 & 219.27 & 424.00 & 305.67 & 585.67 & 346.33 & 472.00 & 383.27 & 801.00 & 647.38 & 652.67 & 700.43 \\
GPT-5 & 744.88 & 378.50 & 560.00 & 310.67 & 1058.17 & 651.67 & 996.70 & 681.67 & 1342.50 & 860.00 & 1167.86 & 958.50 \\
GPT-5-mini & 646.86 & 424.33 & 708.14 & 368.25 & 1079.14 & 704.60 & 1181.27 & 716.50 & 938.33 & 778.00 & 1390.50 & 1061.50 \\
GPT-5-nano & 660.33 & 241.70 & 772.29 & 405.00 & 737.67 & 545.56 & 1135.09 & 537.50 & 706.25 & 832.33 & 1442.38 & 1425.50 \\
\hline
\end{tabular}
}%
\end{table*}

\begin{table*}[htbp]
\centering
\caption{Claude-series Average Token Length by SOP status, correctness, and level}
\label{tab:claude_avg_combined_sop}
\setlength{\tabcolsep}{3pt}
\resizebox{\textwidth}{!}{%
\begin{tabular}{lrrrrrrrrrrrr}
\hline
 & \multicolumn{4}{c}{Level 1} & \multicolumn{4}{c}{Level 2} & \multicolumn{4}{c}{Level 3} \\
Model & \multicolumn{2}{c}{No-SOP} & \multicolumn{2}{c}{SOP} & \multicolumn{2}{c}{No-SOP} & \multicolumn{2}{c}{SOP} & \multicolumn{2}{c}{No-SOP} & \multicolumn{2}{c}{SOP} \\
 & Correct & Incorrect & Correct & Incorrect & Correct & Incorrect & Correct & Incorrect & Correct & Incorrect & Correct & Incorrect \\
\hline
Claude-sonnet-3.5 & 463.00 & 482.62 & 614.80 & 509.88 & 751.50 & 498.70 & 890.40 & 604.12 & 666.00 & 720.75 & 815.33 & 869.83 \\
Claude-sonnet-3.7 & 539.57 & 547.60 & 766.00 & 392.80 & 798.60 & 514.14 & 959.00 & 660.50 & 800.67 & 939.17 & 936.00 & 751.60 \\
Claude-sonnet-4 & 487.00 & 401.20 & 710.67 & 404.88 & 831.25 & 561.88 & 1029.50 & 648.60 & 1014.25 & 811.50 & 1138.50 & 841.50 \\
Claude-opus-4 & 633.00 & 372.62 & 669.00 & 394.33 & 886.20 & 574.43 & 886.86 & 524.50 & 946.00 & 865.75 & 834.33 & 801.86 \\
\hline
\end{tabular}
}%
\end{table*}

\begin{table*}[htbp]
\centering
\caption{Gemini-series Average Token Length by SOP status, correctness, and level}
\label{tab:gemini_avg_combined_sop}
\setlength{\tabcolsep}{3pt}
\resizebox{\textwidth}{!}{%
\begin{tabular}{lrrrrrrrrrrrr}
\hline
 & \multicolumn{4}{c}{Level 1} & \multicolumn{4}{c}{Level 2} & \multicolumn{4}{c}{Level 3} \\
Model & \multicolumn{2}{c}{No-SOP} & \multicolumn{2}{c}{SOP} & \multicolumn{2}{c}{No-SOP} & \multicolumn{2}{c}{SOP} & \multicolumn{2}{c}{No-SOP} & \multicolumn{2}{c}{SOP} \\
 & Correct & Incorrect & Correct & Incorrect & Correct & Incorrect & Correct & Incorrect & Correct & Incorrect & Correct & Incorrect \\
\hline
Gemini-2.5-flash & 224.00 & 143.18 & 409.50 & 251.67 & 523.71 & 316.55 & 728.50 & 386.44 & 801.33 & 549.83 & 774.00 & 616.50 \\
Gemini-2.5-pro & 273.50 & 246.11 & 433.80 & 285.75 & 615.00 & 409.25 & 724.40 & 508.75 & 710.50 & 503.43 & 587.33 & 695.17 \\
\hline
\end{tabular}
}%
\end{table*}

\begin{table*}[htbp]
\centering
\caption{Other models Average Token Length by SOP status, correctness, and level}
\label{tab:other_avg_combined_sop}
\setlength{\tabcolsep}{3pt}
\resizebox{\textwidth}{!}{%
\begin{tabular}{lrrrrrrrrrrrr}
\hline
 & \multicolumn{4}{c}{Level 1} & \multicolumn{4}{c}{Level 2} & \multicolumn{4}{c}{Level 3} \\
Model & \multicolumn{2}{c}{No-SOP} & \multicolumn{2}{c}{SOP} & \multicolumn{2}{c}{No-SOP} & \multicolumn{2}{c}{SOP} & \multicolumn{2}{c}{No-SOP} & \multicolumn{2}{c}{SOP} \\
 & Correct & Incorrect & Correct & Incorrect & Correct & Incorrect & Correct & Incorrect & Correct & Incorrect & Correct & Incorrect \\
\hline
Kimi-K2-Instruct & 395.00 & 461.88 & 710.00 & 587.50 & 1125.50 & 503.80 & 905.00 & 666.43 & 1100.00 & 888.40 & 1530.00 & 1718.67 \\
Deepseek-v3 & 382.00 & 291.45 & 604.67 & 322.12 & 623.31 & 435.75 & 860.17 & 425.14 & 959.50 & 633.12 & 866.00 & 688.50 \\
Qwen3-30b-a3b & 145.00 & 210.20 & 258.00 & 189.08 & 392.19 & 288.55 & 515.00 & 314.92 & 571.00 & 373.50 & 572.31 & 461.22 \\
Qwen3-max & 261.00 & 257.44 & 415.75 & 335.00 & 582.20 & 408.55 & 728.20 & 406.38 & 587.00 & 618.17 & 716.33 & 667.00 \\
\hline
\end{tabular}
}%
\end{table*}

\subsection{Intended use and Compliance}
We generated the dataset via GPT/Claude/Gemini APIs for research and reproducibility, and we do not distribute raw API-generated responses; instead we release prompts, generation scripts/parameters, and evaluation code so others can reproduce results with their own compliant API access.

\subsection{Privacy and Safety Filtering}
We mitigated privacy and safety risks during dataset construction using a two-stage pipeline combining automated screening and manual spot checks. First, we applied a model-based filter to flag samples that may contain personally identifiable information (PII; e.g., names, contact details, addresses) or offensive/harmful content (e.g., hate, harassment, sexual content, or incitement to violence), and we removed flagged samples or redacted sensitive spans with placeholders. Second, we conducted manual audit on randomly sampled subsets after each generation/cleaning iteration to estimate false positives/negatives and iteratively refine filtering thresholds and data-generation constraints to further reduce residual risk. (PII detection and redaction are standard approaches for protecting sensitive information.)

\subsection{Instructions Given to Participants}
We provided all human annotators with a written instruction sheet describing the task goal, step-by-step annotation/verification procedures, and the acceptance criteria used in our human-in-the-loop data pipeline. Annotators were instructed to (i) verify each candidate item, (ii) correct unclear wording while preserving the original meaning, and (iii) reject any item with unverifiable claims, ambiguity, or inconsistencies; only items receiving unanimous approval were kept. The instruction sheet also included an ethics and safety notice: the task involves reading operational documents and editing QA items, poses minimal risk, and annotators should not enter or disclose any personal data; any accidental exposure of sensitive information should be reported and the content should be excluded from the dataset.

\subsection{Characteristics of Annotators}
All data checks and edits were performed by in-house employees. These annotators acted as reviewers who verified and corrected the dataset according to a shared set of guidelines, rather than generating new content. Annotators were selected because of their strong, practical understanding of supply-chain operations. Due to privacy considerations and internal company policy, no individual-level demographic or geographic information was collected, and thus such characteristics cannot be reported or analyzed.

\section{Details of Experiment Settings}
\subsection{Model Size And Budget}
For model size, we evaluate both proprietary API models and models whose specifications are publicly available. For proprietary models (e.g., Claude, Gemini, and Azure OpenAI offerings), vendors do not release exact parameter counts or full training details; therefore, we report only the model names and version identifiers and do not make parameter-size claims for them. For models with publicly documented specifications (e.g., some Qwen/DeepSeek models), we describe their sizes based on official releases; Other models are reported by their publicly available size/spec when provided by the vendor.

For budget, all experiments are conducted via paid inference APIs, and we report the actual billed cost (USD) for running the experiments for each model: Claude Opus 4 (anthropic.claude-opus-4-20250514-v1:0) \(153.0417\); Claude Sonnet 4 (anthropic.claude-sonnet-4-20250514-v1:0) \(33.6668\); Claude 3.7 Sonnet \(22.8242\); Claude 3.5 Sonnet \(22.2992\). Gemini 2.5 Pro \(23.3125\); Gemini 2.5 Flash \(2.1061\). Azure OpenAI: GPT-5 (gpt-5-2025-08-07-GlobalStandard) \(19.2745\); o3-mini (o3-mini-2025-01-31) \(6.0689\); GPT-4o (gpt-4o-2024-11-20) \(4.9204\); GPT-4.1 (gpt-4.1-2025-04-14-GlobalStandard) \(3.0763\); GPT-5-mini (gpt-5-mini-2025-08-07-GlobalStandard) \(1.9212\); GPT-4.1-mini (gpt-4.1-mini-2025-04-14-GlobalStandard) \(1.5056\); o4-mini (o4-mini-2025-04-16-GlobalStandard) \(0.6576\); GPT-5-nano (gpt-5-nano-2025-08-07-GlobalStandard) \(0.6203\). Bailian/others: Qwen3-Max \(1.5692\); Moonshot-Kimi-K2-Instruct \(1.0255\); Qwen3-30B-A3B \(0.2414\); DeepSeek-V3 \(0.7579\). These costs include the runs reported in the paper (including necessary retries).

\subsection{Packages and Parameter Settings}
Our preprocessing, prompting, and evaluation pipeline is primarily implemented with custom scripts rather than third-party NLP toolkits (e.g., NLTK, SpaCy, ROUGE) for tokenization/normalization/metric computation. For answer extraction, models are instructed to produce outputs in a fixed format, and we use regular-expression matching to parse the final predicted label from the generated text. In the context-augmented setting, if a background document exceeds 2,000 characters, we truncate it to the first 2,000 characters and prepend it to the user prompt for consistency under input-length constraints. To support reproducibility, the released code/configuration will record the software environment and dependency versions (e.g., Python version and any SDK/client libraries used for model invocation), along with any decoding settings and random seeds when applicable.

\section{Additional Experiments/Analysis}
\label{sec:additional_experiments}
\subsection{LLM-as-a-Judge Evaluation}
While \textbf{Information Retrieval Accuracy} evaluates whether the model can successfully retrieve the required evidence through tool use, it does not fully capture end-to-end performance, since reasoning errors may still occur after successful tool execution. To address this limitation, we further evaluate the \emph{final outputs} using an LLM-as-a-judge protocol, which serves as a complementary assessment of final-answer correctness. As shown in Table~\ref{tab:tool_calling_only}, the overall trends are largely consistent with those observed under information retrieval accuracy, with only a few models showing slight declines. This suggests that, although some models can retrieve the correct information, they may still omit or distort specific details when synthesizing the final response.
\FloatBarrier
\begin{table}[H]
\caption{Tool-calling performance under SOP-free and SOP-guided settings, evaluated by LLM-as-a-judge. Green values indicate gains, and red values indicate drops.}
\centering
\footnotesize
\setlength{\tabcolsep}{4pt}
\renewcommand{\arraystretch}{0.95}
\begin{tabular}{lcc}
\toprule
\multirow{2}{*}{\textbf{Model}} & \multicolumn{2}{c}{\textbf{Tool Calling Accuracy (\%)}} \\
\cmidrule(lr){2-3}
& w/o SOP & w/ SOP \\
\midrule
\textbf{\textit{Proprietary LLM}} & & \\
GPT-4o             & 20.40 & 16.32 \loss{4.08} \\
GPT-4.1-mini       & 10.39 & 28.57 \gain{18.18} \\
GPT-4.1            & 11.22 & 16.32 \gain{5.11} \\
GPT-O3-mini        & 6.10  & 11.22 \gain{5.11} \\
GPT-5-nano         & 22.03 & 83.21 \gain{61.18} \\
GPT-5-mini         & 42.69 & 86.73 \gain{44.04} \\
GPT-5              & 35.71 & 84.69 \gain{48.98} \\
Claude-3.5-Sonnet  & 11.22 & 29.59 \gain{18.37} \\
Claude-3.7-Sonnet  & 35.71 & 39.09 \gain{3.38} \\
Claude-4-Sonnet    & 31.63 & 39.79 \gain{8.16} \\
Claude-4-Opus      & 23.71 & 34.69 \gain{10.98} \\
Gemini-2.5-Flash   & 8.16  & 23.46 \gain{15.30} \\
Gemini-2.5-Pro     & 11.22 & 37.56 \gain{26.34} \\
\midrule
\textbf{\textit{Open-Source LLM}} & & \\
Deepseek-V3       & 10.20 & 31.63 \gain{21.43} \\
Deepseek-R1       & 17.14 & 18.19 \gain{1.05} \\
Kimi-K2-Instruct  & 12.24 & 10.20 \loss{2.04} \\
Qwen3-Max         & 7.14  & 28.57 \gain{21.43} \\
Qwen3-30B-A3B     & 3.10  & 8.16 \gain{5.06} \\
QwQ-32B           & 5.02  & 10.24 \gain{5.22} \\
\bottomrule
\end{tabular}
\label{tab:tool_calling_only}
\end{table}
\FloatBarrier

\subsection{Controlled Comparison with ReAct}
To ensure a fair comparison, we conduct two additional controlled experiments that separate the effect of the SOP itself from the effect of the ReAct framework and the available computation budget. Specifically, the SOP baseline is evaluated using Pass@5, which matches the baseline-equivalent sampling budget, while Single-Path ReAct is restricted to a single trajectory with at most 10 thought--action steps. As shown in Table~\ref{tab:react_controlled_comparison}, Single-Path ReAct outperforms the SOP (Pass@5) baseline on nearly all evaluated models, with especially large gains for Claude-4-Sonnet, Claude-4-Opus, and Gemini-2.5-Pro. These results suggest that the observed improvements cannot be attributed solely to increased parallel sampling. Instead, they indicate that the ReAct framework itself provides additional reasoning and adaptation capacity in complex supply chain scenarios.
\FloatBarrier
\begin{table}[H]
\caption{Controlled comparison between SOP and ReAct under matched budget settings. The SOP baseline is evaluated with Pass@5, while Single-Path ReAct is limited to a single trajectory with at most 10 thought--action steps.}
\centering
\footnotesize
\setlength{\tabcolsep}{4pt}
\renewcommand{\arraystretch}{0.95}
\begin{tabular}{lcc}
\toprule
\textbf{Model} & \textbf{SOP (Pass@5)} & \textbf{Single-Path ReAct} \\
\midrule
GPT-4o            & 18.36 & 31.63 \gain{13.27} \\
GPT-4.1-mini      & 31.63 & 33.96 \gain{2.33} \\
GPT-4.1           & 18.36 & 18.31 \loss{0.05} \\
Claude-3.5-Sonnet & 37.76 & 38.77 \gain{1.01} \\
Claude-4-Sonnet   & 40.82 & 71.42 \gain{30.60} \\
Claude-4-Opus     & 42.85 & 70.43 \gain{27.58} \\
Gemini-2.5-Pro    & 37.76 & 70.53 \gain{32.77} \\
Qwen3-Max         & 30.61 & 31.71 \gain{1.10} \\
Qwen3-30B-A3B     & 11.22 & 19.38 \gain{8.16} \\
\bottomrule
\end{tabular}
\label{tab:react_controlled_comparison}
\end{table}
\FloatBarrier

\section{Prompt in Detail}
Figures~\ref{fig:qg-prompt}, \ref{fig:qg-prompt_continued}, \ref{fig:analysis-prompt}, and \ref{fig:final-prompt} are prompts used in this work.

\onecolumn





\begin{figure}[p] 
\centering
\begin{tcolorbox}[
  enhanced,
  breakable,
  colback=white,
  colframe=black,
  boxrule=0.5pt,
  left=6pt,right=6pt,top=6pt,bottom=6pt
]
\begin{Verbatim}[breaklines,breakanywhere, fontsize=\small]
You are a professional question generation master. Your goal is to generate a multiple-select question from a given paragraph. The content of the multiple-select question should be based on the input text without introducing external content. Centered on the user-specified concept {TERM}, identify core entities, terms, definitions, or logical relationships about the concept {TERM} from the user-provided reference text, and strictly follow the requirements below to generate a single-choice question:

<requirement>

# Requirements for the output format:
The output result is a JSON object, specifically as follows
<example>
{"question":"Stem",
"options":[{"key":"A","text":"Option A"},{"key":"B","text":"Option B"},{"key":"C","text":"Option C"},{"key":"D","text":"Option D"}],"                                                    
"answer":["List of correct option letters"],
"question_type":"multiple_choices",                                                   
"explanation":"Explanation content",
"type":"Main category",
"sub_type":"Sub-category",
"question_type":"multiple_choices",
"tags":["Concept 1","Concept 2","Concept 3"]}
</example>

The output result includes:

## Core question content
1. "question": The stem, a string that contains the specific description of the question. The stem should raise a clear multiple-select question around the key concept (e.g., “Which of the following are...?”). The stem should be clearly expressed and concise, and should not include phrases like “according to the article.” Ask directly. Avoid overly absolute statements in the stem.
2. "options": Options, an array providing 4 options (A - D), with 1-4 correct answers and the rest being reasonable distractors. Each option is a dictionary with two fields, "key" and "text". The "key" field is the identifier for the option, and the "text" field is the textual description of the option. For example, {"key":"A","text":"Option A"}.
3. "answer": The answer, an array containing the strings of the correct answers, i.e., multiple of the 4 options (A - D). For example: "answer": ["A","B"].
4. "explanation": Explanation, a string providing a detailed explanation for the correct answers, explaining why the option is correct. The explanation should state the reasons for the correct options and common misunderstandings for the incorrect options.

## Classification and metadata
1. "field": A string that defines the macro field or theme to which the question belongs. It can only be one of the following:
   1. Fulfillment, Expression, Procurement, Returns and Waste, In-warehouse, Inventory, Merchandise, Merchant, Network, Planning, Finance and Operations, Permissions & Accounts, Logistics Collaboration, Other
2. "sub_field": A string that provides a more detailed division of the main category.
3. "question_type": A string that explicitly indicates the form of the question as "multiple_choices".
4. "tags": An array of strings that provides a series of keywords related to the content of the question for searching and filtering.

## Difficulty definition
High-difficulty questions usually have one or more of the following characteristics:
1. Applying: Applying learned knowledge to new scenarios or solving practical problems.
2. Analyzing: Breaking information into different parts and exploring their relationships and structures. The question usually contains irrelevant information that needs to be filtered.
3. Synthesizing: Answering requires a full understanding of the text and some reasoning ability to arrive at the answer.
\end{Verbatim}
\end{tcolorbox}
\caption{Question generation prompt (part 1 of 2).}
\label{fig:qg-prompt}
\end{figure}

\clearpage 

\begin{figure}[p]
\centering
\begin{tcolorbox}[
  enhanced,
  breakable,
  colback=white,
  colframe=black,
  boxrule=0.5pt,
  left=6pt,right=6pt,top=6pt,bottom=6pt
]
\begin{Verbatim}[breaklines,breakanywhere, fontsize=\small]
## Output format validation 
All fields must be wrapped in double quotes, the JSON structure must be complete (e.g., brackets closed), and the output must be directly parseable as a JSON object. Do not add ```json or other extra characters.
</requirement>

# Output example:
<example>
{
"question":"Which of the following are common payment methods in medical expense payments? (Multiple select)",                                                    
"options":[{"key":"A","text":"Fee-for-service"},
{"key":"B","text":"Capitation"},                          
{"key":"C","text":"Wage-based payment"},
{"key":"D","text":"Diagnosis-related group (DRG) payment"}],                                  
"answer":["A","B","D"],
"question_type":"multiple_choices", 
"explanation":"Fee-for-service (A) is a model that pays based on the quantity of services provided. Capitation (B) pays a fixed amount per enrolled beneficiary, regardless of service utilization. DRG (D) is a prospective payment system based on diagnostic categories. Wage-based payment (C) is not a standard medical reimbursement model.",
"field":"Healthcare",   
"sub_field":"Medical payment methods",   
"tags":["Healthcare","Medical payment methods","Capitation","Fee-for-service","DRG"]
}
</example>

# Input reference text:
<input>
{TEXT}
</input>
\end{Verbatim}
\end{tcolorbox}
\caption{Question generation prompt (continued).}
\label{fig:qg-prompt_continued}
\end{figure}

\clearpage


\begin{tcolorbox}[
  enhanced,
  breakable,          
  colback=white,
  colframe=black,
  boxrule=0.5pt,
  left=6pt,right=6pt,top=6pt,bottom=6pt
]
\begin{Verbatim}[breaklines,breakanywhere, fontsize=\small]
Please analyze the following data generation result and propose specific revision suggestions to improve quality:

Original prompt:
<original_prompt>
{REPLACE_ORIGINAL_PROMPT}
</original_prompt>

Generated result:
<initial_result>
{REPLACE_INITIAL_RESULT}
</initial_result>
Please provide specific and accurate revision suggestions.

\end{Verbatim}

\end{tcolorbox}
\captionof{figure}{Adversarial Review and Refinement.}
\label{fig:analysis-prompt}
\begin{tcolorbox}[
  enhanced,
  breakable,          
  colback=white,
  colframe=black,
  boxrule=0.5pt,
  left=6pt,right=6pt,top=6pt,bottom=6pt
]
\begin{Verbatim}[breaklines,breakanywhere, fontsize=\small]
Based on the following analysis suggestions, regenerate the improved data:

Original prompt:
<original_prompt>
{REPLACE_ORIGINAL_PROMPT}
</original_prompt>

Initial generated result:
<initial_result>
{REPLACE_INITIAL_RESULT}
</initial_result>

Modification suggestions:
<modification_suggestions>
{REPLACE_MODIFICATION_SUGGESTIONS}
</modification_suggestions>

Note that the returned result must meet the requirements of the original prompt. Do not add additional modification notes; directly return the regenerated improved result only!

\end{Verbatim}

\end{tcolorbox}
\captionof{figure}{Final Synthesis Prompt.}
\label{fig:final-prompt}


\end{document}